\documentclass{article}

\usepackage[a4paper, margin=1in]{geometry} % 替换fullpage
\usepackage[utf8]{inputenc}
\usepackage[T1]{fontenc}
\usepackage{microtype}
\usepackage{indentfirst}

\usepackage{amsmath}
\usepackage{amssymb}
\usepackage{amsfonts}
\usepackage{amsthm}
\usepackage{mathtools}
\usepackage{bm}
\usepackage{commath}
\usepackage{relsize}
\usepackage{bbm}
\usepackage{nicefrac}
\usepackage{mysymbol}

\usepackage{graphicx}
\usepackage{tikz}
\usetikzlibrary{positioning}
\usepackage{pgfplots}
\pgfplotsset{compat=1.18}
\usepackage{subcaption} % 替换subfigure
\usepackage{multirow}
\usepackage{diagbox}
\usepackage{booktabs}
\usepackage{hhline}
\usepackage{array}
\usepackage{float}

\usepackage{algorithm}
\usepackage{algorithmic}
\usepackage{listings}

\usepackage{cite}

\usepackage{pifont}
\usepackage{enumerate}
\usepackage{enumitem}
\usepackage[title]{appendix}
\usepackage[font={small}]{caption}

\newtheorem{theorem}{Theorem}[section]
\newtheorem{proposition}[theorem]{Proposition}

\theoremstyle{definition}

\theoremstyle{remark}
\newtheorem{remark}[theorem]{Remark}

\allowdisplaybreaks
\newcolumntype{?}{!{\vrule width 1pt}}

\makeatletter
\def\Let@{\def\\{\notag\math@cr}}
\makeatother

\usepackage[hidelinks,backref=page]{hyperref}
\definecolor{darkred}{RGB}{150,0,0}
\definecolor{darkgreen}{RGB}{0,150,0}
\definecolor{darkblue}{RGB}{0,0,150}
\hypersetup{colorlinks=true, linkcolor=darkred, citecolor=darkblue, urlcolor=darkblue}

\renewcommand*{\backref}[1]{}
\renewcommand*{\backrefalt}[4]{%
    \ifcase #1 (Not cited.)%
    \or        (Cited on page~#2.)%
    \else      (Cited on pages~#2.)%
    \fi}
    
\title{\textbf{pFedFair: Towards Optimal Group Fairness-Accuracy Trade-off in
Heterogeneous Federated Learning}}

\date{}

\author{
Haoyu LEI\thanks{Department of Computer Science and Engineering, The Chinese University of Hong Kong, hylei22@cse.cuhk.edu.hk},
Shizhan Gong\thanks{Department of Computer Science and Engineering, The Chinese University of Hong Kong, szgong22@cse.cuhk.edu.hk},
Qi Dou\thanks{Department of Computer Science and Engineering, The Chinese University of Hong Kong, qdou@cse.cuhk.edu.hk},
Farzan~Farnia\thanks{Department of Computer Science and Engineering, The Chinese University of Hong Kong, farnia@cse.cuhk.edu.hk}
	}

\begin{document}
\maketitle

\begin{abstract}
Federated learning (FL) algorithms commonly aim to maximize clients’ accuracy by training a model on their collective data. However, in several FL applications, the model’s decisions should meet a group fairness constraint to be independent of sensitive attributes such as gender or race. While such group fairness constraints can be incorporated into the objective function of the FL optimization problem, in this work, we show that such an approach would lead to suboptimal classification accuracy in an FL setting with heterogeneous client distributions. To achieve an optimal accuracy-group fairness trade-off, we propose the Personalized Federated Learning for Client-Level Group Fairness (pFedFair) framework, where clients locally impose their fairness constraints over the distributed training process. Leveraging the image embedding models, we extend the application of pFedFair to computer vision settings, where we numerically show that pFedFair achieves an optimal group fairness-accuracy trade-off in heterogeneous FL settings. We present the results of several numerical experiments on benchmark and synthetic datasets, which highlight the suboptimality of non-personalized FL algorithms and the improvements made by the pFedFair method.    

\end{abstract}

%\vspace{-.1cm}
\section{Introduction}

\begin{figure}[t]
    \centering
        \includegraphics[width=0.85\textwidth]{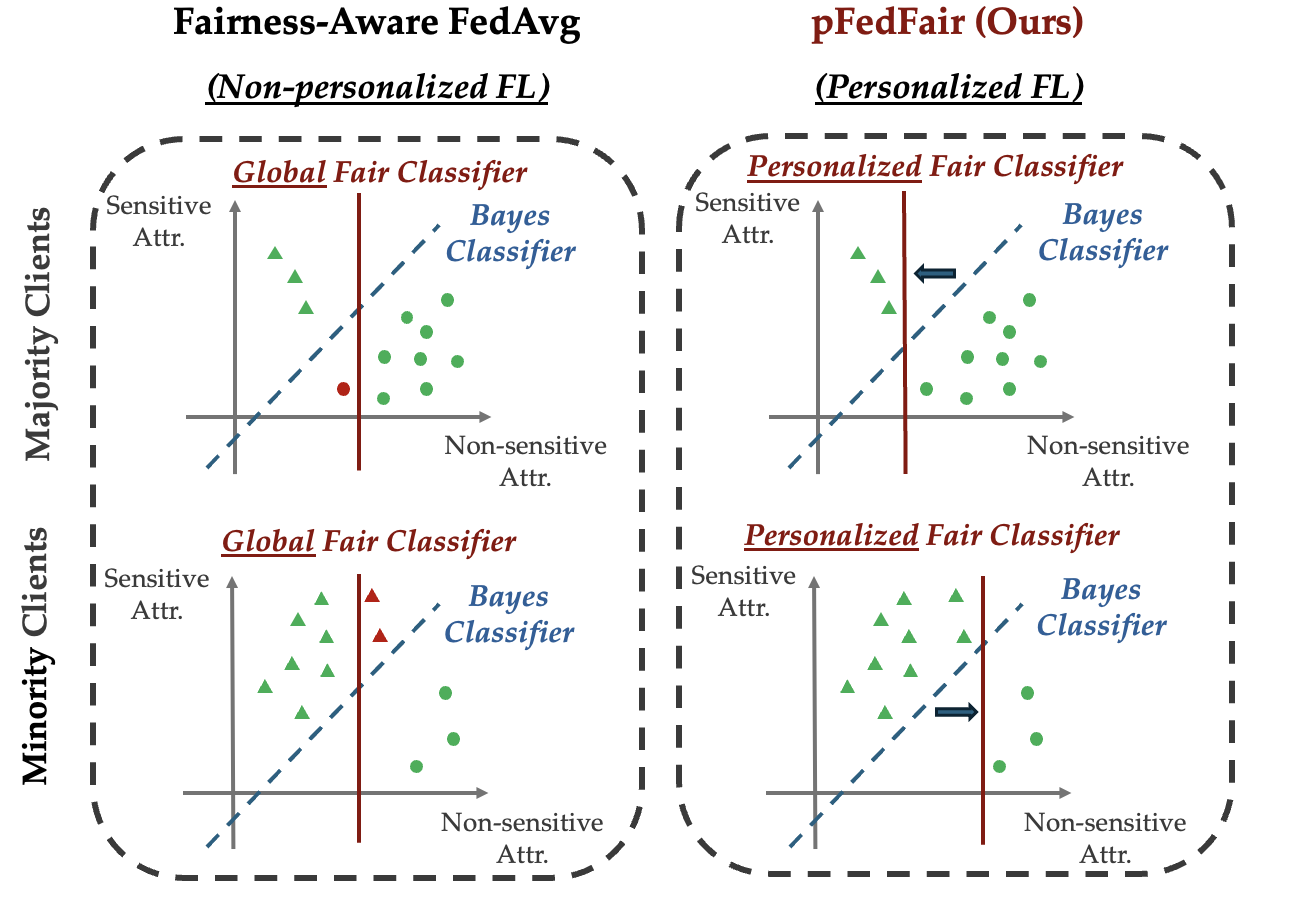}%
    \caption{
    Illustration of client-level fairness-aware personalization in heterogeneous federated learning. The fairness-aware global model aggregation often favors majority sensitive groups due to client heterogeneity. Our proposed \textbf{pFedFair} enables each client to learn a personalized fair classifier, optimizing the client-level fairness-accuracy trade-off while using global model knowledge.
    }
    \label{fig:figure1_1}
\end{figure}

The increasing deployment of supervised learning models in high-stakes decision-making domains, such as hiring, lending, and criminal justice, necessitates a rigorous analysis of their potential for perpetuating and amplifying societal biases. Specifically, undesired correlations between model predictions and sensitive attributes (e.g., race) present a significant challenge, which has motivated substantial research in fairness-aware machine learning~\cite{barocas2019fairness, mehrabi2021survey}. Within this area, \textit{group fairness}, which aims to ensure statistical parity in predictions across protected groups defined by sensitive attributes, has received considerable attention~\cite{dwork2012fairness, zemel2013learning, hardt2016equality}. While many methods have been proposed for achieving various notions of group fairness (e.g., demographic parity~\cite{calders2009building}, equal opportunity, and equalized odds~\cite{hardt2016equality}) in \textit{centralized} learning settings, their applicability and effectiveness in \textit{distributed} settings, particularly \textit{federated learning (FL)}~\cite{mcmahan2017communication}, remain relatively underexplored.

Existing centralized fairness-aware algorithms often leverage regularization techniques, incorporating penalty terms into the objective function that explicitly discourage statistical dependence between model predictions and sensitive attributes. These penalties are typically based on measures of dependence, such as mutual information~\cite{kamishima2012fairness}, Pearson correlation~\cite{zafar2017fairness}, or the Hirschfeld-Gebelein-R\'enyi (HGR) maximal correlation~\cite{mary2019fairpca, baharlouei2019r}. These methods have demonstrated empirical success in mitigating bias while maintaining acceptable predictive accuracy. 

However, the inherent characteristics of FL, notably \textit{client heterogeneity}~\cite{li2020federated, kairouz2021advances}, pose significant challenges to the direct application of these centralized techniques for boosting group fairness. In standard FL, multiple clients, each possessing a local dataset, collaboratively train a shared global model without explicitly exchanging their data. This collaborative paradigm offers the potential for improved performance by leveraging a larger yet distributed dataset. Nevertheless, heterogeneity in client data distributions, a common characteristic of real-world FL deployments, can induce substantial variations in model performance across clients~\cite{zhao2018federated}. Specifically, when heterogeneity affects the distribution of sensitive attributes, achieving client-level group fairness can become a challenging task.

The presence of heterogeneous sensitive attribute distributions across clients leads to the following conflict: the notion of a globally optimal fair classifier would no longer apply as it is used in the centralized case. For example, consider a multi-branch corporation where applicant demographics vary significantly across branches; one branch may have a high proportion of male applicants, while another may have a higher fraction of female applicants. Under such a scenario, enforcing group fairness at the individual branch level can necessitate different client-level decision-making models, thereby undermining the assumption of a single globally shared model in standard FL algorithms.

In this work, we introduce the \textit{Personalized Federated Learning for Client-Level Group Fairness (pFedFair)} framework, specifically designed to address client-level group fairness constraints in heterogeneous FL environments, shown in Figure \ref{fig:figure1_1}. The pFedFair approach adopts a \textit{targeted personalization} strategy, where the model personalization focuses on the \textit{fairness constraint}, rather than the entire supervised learning setting. Our numerical and theoretical results indicate the potential suboptimality of globally-learned models (e.g., FedAvg~\cite{mcmahan2017communication}, FedProx~\cite{li2020federated}) and fully-personalized models (e.g., pFedMe~\cite{dinh2020personalized}, Per-FedAvg~\cite{fallah2020personalized}) in settings where the underlying Bayes optimal classifier (in the absence of fairness constraints) is shared across clients, but the distributions of sensitive attributes exhibit heterogeneity. This setting is of practical relevance, as data heterogeneity in FL deployments often impacts the marginal distribution of features, \(P(X)\), while the conditional label distribution, \(P(Y|X)\), remains consistent across clients~\cite{yurochkin2019bayesian}.

Our empirical results supports the theoretical discussion on federated learning under client-level group fairness constraints, indicating an improved performance and better generalization capability of pFedFair in comparison to both baseline and fully personalized FL algorithms when \(P(Y|X)\) is shared and sensitive attribute distributions within \(P(X)\) are heterogeneous. The enhanced generalization is a consequence of the targeted application of personalization, focusing solely on the fairness component.

Furthermore, we leverage image embedding models to apply the proposed group fairness regularization to standard computer vision datasets. To do this, we employ the widely-used CLIP~\cite{radford2021learning} and  DINOv2~\cite{oquab2023dinov2} models. Our numerical experiments suggest that both in the centralized and federated learning settings, a simple linear model trained over the embedding can achieve an improved group fairness-accuracy trade-off compared to the direct application of the group-fairness regularization to standard ResNet~\cite{he2016deep}, ViT~\cite{dosovitskiy2020image}, and CNN image classifiers. Furthermore, the application of image embedding enables efficient group-fairness-constrained federated training via the pFedFair method. Our numerical analysis over the real CelebA~\cite{liu2018large}, UTKFace~\cite{zhang2017age}, and synthetic image datasets with text-to-image models support applying pFedFair to image-classification problems with group-fairness constraints. The main contributions of this work are summarized as follows:
\begin{itemize}[leftmargin=*]
    \item Analyzing the heterogeneous FL setting with client-level group fairness constraints
    \item Proposing the pFedFair to address personalized federated learning under heterogeneous feature distribution
    \item Proposing the application of image embeddings to improve the performance of group-fairness learning methods in FL on computer vision datasets 
    \item Presenting numerical results supporting the application of personalized FL to the sensitive attribute variable under group fairness constraints.
\end{itemize}

\section{Related Works}

\textbf{Fairness in Centralized Machine Learning.} studies on group fairness aims to minimize the dependence of algorithmic decisions on biased demographic groups. A common strategy involves adding fairness regularization terms to the learning objective, addressing fairness notions like Demographic Parity (DP) and Equalized Odds (EO), as studied in~\cite{dwork2012fairness, hardt2016equality}. In centralized settings, various methods have been proposed to ensure group fairness, including fairness-aware optimization~\cite{kamishima2011fairness, zafar2017fairness, zhang2018mitigating}, reweighting~\cite{rezaei2020fairness, cho2020afair, beutel2019putting}, and adversarial approaches~\cite{mary2019fairness, baharlouei2019r, grari2021learning}. For visual recognition, recent efforts focus on fairness regularization~\cite{hanel2022enhancing, sagawa2019distributionally}, reweighting data distributions~\cite{kim2023fair, lv2023duet}, and using generative models to synthesize diverse training samples~\cite{li2023bias, ramaswamy2021fair, peychev2022latent}. These approaches address inherent biases while improving fairness performance across tasks.

\noindent\textbf{Fairness in Federated Learning.} Fairness in FL~\cite{mcmahan2017communication} introduces unique challenges compared to centralized settings. Among fairness perspectives in FL~\cite{shi2023towards}, \emph{client fairness}~\cite{li2019fair} focuses on equitable performance across clients, while our work addresses \emph{group fairness}, targeting fairness for protected subgroups, such as gender or race. A key distinction in FL is between \emph{global fairness}, which ensures fairness across all clients, and \emph{local fairness}, which addresses fairness within individual clients’ datasets. Most existing methods emphasize global fairness, such as aggregating locally fair models~\cite{chu2021fedfair, galvez2021enforcing, zhang2020fairfl} or re-weighting techniques~\cite{ezzeldin2023fairfed, abay2020mitigating, du2021fairness, zeng2021improving}. However, global fairness does not guarantee local fairness, as client-specific sensitive attributes may vary. Research on local fairness remains limited, with notable contributions including constrained optimization approaches~\cite{cui2021addressing} and tradeoff analyses between global and local fairness~\cite{hamman2023demystifying}. We aim to improve local fairness, making the model truly deployable at the client level.

\noindent\textbf{Personalized Federated Learning.} 
Personalized Federated Learning (PFL) aims to address the challenge of heterogeneity in user data distributions under the FL setting~\cite{tan2022towards}. Several approaches have been proposed for the goal, including meta-learning~\cite{t2020personalized,fallah2020personalized}, multi-task learning~\cite{smith2017federated,li2021ditto}, model parameters decomposition~\cite{collins2021exploiting,oh2021fedbabu}, model mixture~\cite{hanzely2020federated,hanzely2020lower}, parameter fine-tuning~\cite{deng2024unlocking,tamirisa2024fedselect}, adaptive learning rate~\cite{li2021fedbn,jiang2024heterogeneous}, and model editing~\cite{yuan2024pfededit}. These methods demonstrate the efficacy regarding addressing the challenge of local clients' heterogeneity. However, they primarily focus on performance optimization without explicitly addressing fairness concerns. On the other hand, our work aims to improve client-level group fairness through the use of PFL, which remains largely unexplored.

\section{Preliminaries}

\subsection{Group Fairness in Machine Learning}
Consider a classification setting where the goal is to predict a label $y\in\mathcal{Y}$ from an observation of a feature $x\in\mathcal{X}$. In supervised learning, the learner observes $n$ labeled samples $(x_i,y_i)_{i=1}^n$ drawn from a distribution $P_{X,Y}$ and aims to use these samples to find a classifier $f\in\mathcal{F}$ that predicts the label $y\approx f(x)$ for an unseen test sample $(x,y)$ also drawn from $P_{X,Y}$. A standard approach to this classification problem is empirical risk minimization (ERM), where, for a given loss function $\ell : \mathcal{Y}\times\mathcal{Y}\rightarrow\mathbb{R}$, we aim to solve
\begin{equation}\label{Eq: ERM Vanilla}
    \min_{f\in\mathcal{F}}\: \widehat{\mathcal{L}}(f) := \frac{1}{n}\sum_{i=1}^n \ell\bigl(f(x_i) , y_i\bigr)
\end{equation}
Note the empirical risk function $\widehat{\mathcal{L}}(f)$ estimates the population risk $\mathcal{L}(f)=\mathbb{E}\bigl[\ell(f(X),Y)\bigr]$ that is the expected value according to the underlying distribution of test samples.

The group fairness constraint considers a sensitive attribute variable $S\in\mathcal{S}$, e.g., the gender or race feature, and aims to find a decision rule in $\mathcal{F}$ that leads to an $S$-independent decision. According to the standard Demographic Parity (DP), the learner aims for a statistically independent prediction $f(X)$ of a sensitive attribute $S$. To quantify the lack of demographic parity, a standard measure is the Difference of Demographic Parity (DDP), which is defined for a binary sensitive attribute $S\in\{0,1\}$ as:
\begin{align}\label{Eq: DDP Definition}
    \mathrm{DDP}(f):= \sum_{y\in\mathcal{Y}}\:\Bigl\vert &P\bigl(f(X)=y\big|S=0\bigr)\nonumber
    - P\bigl(f(X)=y\big|S=1\bigr)  \Bigr\vert
\end{align}
To enforce DP in the supervised learning, a common approach is to incorporate a new term in the loss function $\rho(f(X),S)$, that measures the dependence of the random variables $f(X)$ and $S$, which leads to the following optimization problem given regularization parameter $\eta \ge 0$:   
\begin{equation}\label{Eq: ERM with Fairness}
    \min_{f\in\mathcal{F}}\:\Bigl\{ \widehat{\mathcal{L}}_{\text{\rm fair}}(f)\, :=\, \widehat{\mathcal{L}}(f) + \eta\cdot\rho(f(X),S) \Bigr\}.
\end{equation}
Standard choices of dependence measure $\rho$ include the DDP~\cite{cho2020bfair}, the mutual information~\cite{cho2020afair}, Pearson correlation~\cite{beutel2019putting}, the HGR maximal correlation~\cite{mary2019fairness}, and the exponential Renyi mutual information (ERMI)~\cite{lowy2022stochastic}.
\subsection{Personalized Federated Learning}
Federated Learning (FL) aims to train a decentralized model through the private cooperation of multiple clients connected to a central server. The standard formulation of the FL problem for training a single global model $f\in\mathcal{F}$ over a network with $m$ clients is given by:
\begin{equation}\label{Eq: FL problem}
    \min_{f\in\mathcal{F}} \: \frac{1}{m}\sum_{i=1}^m \widehat{\mathcal{L}}_i(f), 
\end{equation}
where $\widehat{\mathcal{L}}_i(f)$ denotes the empirical risk of Client $i$. In other words, standard (non-personalized) FL algorithms (e.g., FedAvg) aim to minimizes the average risk across clients.

PFL focuses on the case of non-IID data across clients, where the distributions of data at different clients do not necessarily match. Standard formulations of PFL seek to find a global model $f\in\mathcal{F}$ whose neighborhood contains functions that perform optimally for each client. For example, the pFedMe~\cite{t2020personalized} algorithm minimizes the averaged Moreau envelope of the loss functions:
\begin{equation}\label{Eq: PFL problem}
    \min_{f\in\mathcal{F}} \: \frac{1}{m}\sum_{i=1}^m \Bigl\{\mathcal{L}^{(\gamma)}_i(f):= \min_{f_i \in\mathcal{F}} \mathcal{L}_i(f_i) + \frac{\gamma}{2} \bigl\Vert f_i - f\bigr\Vert^2\Bigr\}. 
\end{equation}
In the above, $\mathcal{L}^{(\gamma)}_i$ denotes the Moreau envelope of Client $i$'s loss function with the Moreau-Yoshida regularization parameter $\gamma>0$. The objective is to find the global model $f$, where each client's personalized model $f_i$, obtained by solving the inner optimization problem, achieves optimal performance for that client.

\section{pFedFair: Personalized Federated Learning for Client-Level Group Fairness}

In this work, we focus on achieving group fairness in a heterogeneous FL setting. In FL, group fairness requires each client to enforce demographic parity within their local dataset, i.e., the DP condition is evaluated independently by each client. A practical example is a corporation with multiple departments, where each department seeks to make fair and accurate hiring decisions that are independent of sensitive attributes. Notably, in an FL setting with heterogeneous data distributions, client-level and global demographic parity can diverge. In this work, we specifically aim for client-level demographic parity.

First, our analysis suggests that using a single global model for all clients can lead to suboptimal classifiers, even when clients share the same conditional distribution $ P(Y|X,S) $, meaning that given the features and sensitive attribute of a sample, the label assignment follows an identical conditional distribution.
\begin{proposition}\label{Thm: 1}
Consider $ m $ clients in an FL task with $ 0/1 $-loss, where $ P^{(i)}(X,S,Y) $ represents the joint distribution of Client $ i\in\{1,\ldots ,m\} $ and binary $S\in\{0,1\}$. Suppose the conditional distribution $ P^{(i)}(Y|X,S) $ is shared across clients, where $Y= g(X,S)$ for a labeling scheme $g$. 
\begin{itemize}
    \item[(a)] Regardless of the consistency of $ P(X,S) $ across clients, all clients share the same optimal Bayes classifier $ g(x,s) = \arg\!\max_y P(y|x,s) $.
    \item[(b)] Assuming demographic parity within each client, i.e., $ f_i(X_i)\perp S_i $ for all $ i $, the optimal decision rule $ f^*_i $ for each Client $ i $ can differ. Specifically, for any $ f\in\mathcal{F} $, the performance gap between the global classifier $ f $ and the optimal client-specific classifiers $ f_i $ satisfies the following, where $ s_{\max},s^{(i)}_{\max} $ denote the majority sensitive attribute outcomes in the network and Client $ i $, respectively,
    \begin{align*}
        &\sum_{i=1}^m \mathcal{L}_i(f) - \mathcal{L}_i(f^*_i) \:
        \ge\: \sum_{i=1}^m \bigl(2P^{(i)}(S= s_{\max})-1\bigr)\times\\
        & \Bigl\vert \sum_{y\in\mathcal{Y}}P^{(i)}(Y=y|S= s_{\max})- P^{(i)}(Y=y|S= s^{(i)}_{\max})\Bigr\vert.  
    \end{align*}
    
\end{itemize}
\end{proposition}
\begin{proof}
We defer the proof to the Appendix.
\end{proof}
\begin{remark}\label{Remark: 1}
The above proposition highlights that while the optimal Bayes classifiers remain the same across clients when they share $ P(y|x,s) $, their optimal classifiers will differ under the demographic parity constraint if the majority sensitive attribute varies between clients. For instance, if the majority gender of applicants differs across hiring departments, their fair classifiers will not match, even if the relationship between applicants' qualifications and their features remains identical.
\end{remark}

\begin{figure*}[t]
    \centering
    \includegraphics[width=0.98\textwidth]{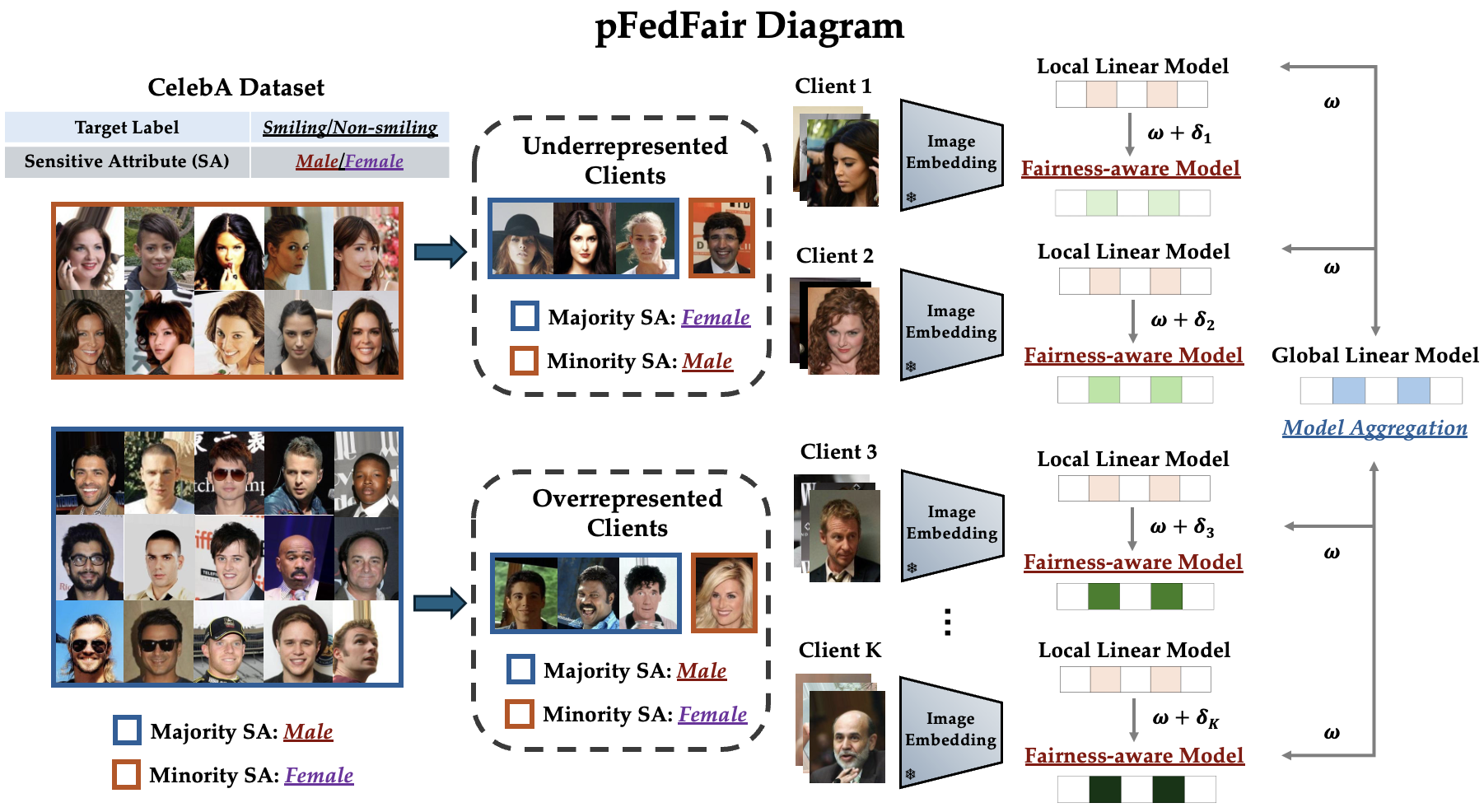}%
    \caption{Overview of \textbf{pFedFair} framework in heterogeneous federated learning settings. The framework integrates a global model optimized for utility with fairness-aware personalization at each client to achieve client-level optimal fairness-accuracy trade-offs.}
    \label{fig:figure1_2}
\end{figure*}

As discussed in Proposition~\ref{Thm: 1} and Remark~\ref{Remark: 1}, enforcing client-level group fairness in federated learning requires personalized classifiers, even when clients share the same labeling scheme. A natural question then arises: How should personalization be incorporated into the FL process? In the following discussion, we argue that an optimal PFL framework should satisfy two key objectives:  
1) The global model should minimize the overall (non-group-fair) loss across all clients.  
2) Each client's personalized model should minimize the local loss while satisfying group fairness constraints.  In other words, an effective PFL scheme should ensure that the global model $ f_w $ generalizes well over all clients' data (similar to FedAvg's objective), while each client’s personalized model $ f_{w_i} $ optimally balances fairness and accuracy.

The following proposition suggests that, under a shared labeling scheme, personalizing the optimal Bayes model achieves near-optimal performance. Consequently, the two objectives outlined above align with the optimality conditions in the group-fair FL task.
\begin{proposition}
Consider the setting of Proposition~\ref{Thm: 1}. Consider divergence measure $d$ is used for the fairness constraint of each Client $i$, i.e., $\rho (\widehat{Y},S)= \mathbb{E}[d(P_{\widehat{Y}|S=s},P_{\widehat{Y}})] \le \epsilon$. Then, the optimal decision rule at Client $i$ will be the solution $Q^{*}_{\widehat{Y}|X,S}$ of the following optimal transport-based minimization problem:
\begin{align}
&\min_{Q_{Y,\widehat{Y}|X,S}}\quad\; \mathbb{E}_{P^{(i)}_{X,S}\times Q_{Y,\widehat{Y}|X,S}}\Bigl[\ell_{0/1}(Y,\widehat{Y})\Bigr]\nonumber\\
&\text{\rm subject to}\;\;\;\; Q_{Y|X,S}=P_{Y|X,S}\\
&\qquad\qquad\;\;\;\: \mathbb{E}_{s\sim P^{(i)}_S}[d(Q_{\widehat{Y}|S=s},Q_{\widehat{Y}})]\le \epsilon. \nonumber
\end{align}
\end{proposition}

\begin{algorithm}[t]
    \caption{pFedFair Fairness-aware Federated Learning} 
    \label{algo:pFedFair}
    \begin{algorithmic}[1]
    \STATE \textbf{Input:}  Data $\{(\mathbf{x}_{i,j},s_{i,j},,y_{i,j})_{j=1}^n\}_{i=1}^m$ of Client $i$,   parameters $\lambda,\eta , \gamma$, fairness measure $\rho$, stepsize $\alpha$\vspace{1mm}
    \STATE \textbf{Initialize} global weight $w$ \vspace{1mm}
        \FOR{$\text{t} \in \{ 1, \ldots , T \}$}\vspace{1mm} 
            \STATE Each Client $i$ computes gradient the clean loss:\vspace{-2mm} 
            $$g^{(i)}_{w} = \frac{1}{n}\sum_{j=1}^n\nabla_{w}\ell\bigl( f_{w}(\mathbf{x}_{i,j}), y_{i,j}\bigr) %+ \lambda \nabla_{w}\rho\bigl(f_{w}(\mathbf{X}),S\bigr)
            $$ \vspace{-3mm}
            \STATE Each Client $i$ computes the weights of the Moreau envelope-personalized fair model:\vspace{-2mm} 
            $$w^{(\gamma)}_i = \underset{w_i}{\arg\!\min}\:  \frac{1}{n}\sum_{j=1}^n\ell\bigl( f_{w_i}(\mathbf{x}_{i,j}), y_{i,j}\bigr) + \frac{\gamma}{2} \bigl\Vert w_i - w\bigr\Vert^2 %+ \lambda \nabla_{w}\rho\bigl(f_{w}(\mathbf{X}),S\bigr)
            $$ \vspace{-3mm}
            \STATE Each Client $i$ computes gradient of Moreau envelope-personalized loss:\vspace{-2mm} 
            $$\hspace{-4mm}g^{(i),\mathrm{fair}}_{w} = \frac{1}{n}\sum_{j=1}^n\nabla_{w}\ell\bigl( f_{w^{(\gamma)}_i}(\mathbf{x}_{i,j}), y_{i,j}\bigr) + \eta\rho\bigl(f_{w^{(\gamma)}_i}(\mathbf{X}),S\bigr)
            $$ \vspace{-3mm}
            \STATE Each Client $i$ updates ${w}$ by gradient descent: $$\widetilde{w}_i \leftarrow {w} - \alpha \bigl(g^{(i)}_{w}+\lambda\,g^{(i),\mathrm{fair}}_{w}\big)$$\vspace{-3mm} 
            \STATE Server synchronizes the weights: $w\leftarrow \frac{1}{m}\sum_{i=1}^m\widetilde{w}_i$\vspace{1mm}
        \ENDFOR\vspace{1mm}
        \STATE \textbf{Output:}  Global model weight $w$ and the clients' personalized model weights $w_1^{(\gamma)},\ldots ,w_m^{(\gamma)}$
    \end{algorithmic}
\end{algorithm}

\begin{proof}
We defer the proof to the Appendix.
\end{proof}
\begin{remark}
As Proposition~2 notes, the optimal fair prediction rule $ f_i^* $, corresponding to the optimal conditional distribution $ Q^{*,(i)}_{\widehat{Y}|X,S} $, is obtained by personalizing the globally optimal $ P_{Y|X,S} $ through solving the optimal transport problem in Proposition~2. This result suggests that the globally trained model $ f $ can correspond to $ P_{Y|X,S} $ to achieve low (non-fair) prediction loss, while each personalized model $ Q^{*,(i)}_{\widehat{Y}|X,S} $ should optimize for group fairness on Client $ i $'s data.
\end{remark}

Therefore, we propose the \emph{personalized pFedFair algorithm} (illustrated in Fig.~\ref{fig:figure1_2}), aiming to satisfy two key objectives:  
1) Minimizing the overall prediction loss via the global model.  
2) Minimizing the group fairness loss via each client’s personalized model. This leads to the following optimization formulation with a regularization parameter $ \lambda>0 $:
\begin{align}\label{Eq: PFedFair problem}
    &\min_{f\in\mathcal{F}}  \frac{1}{m}\sum_{i=1}^m \Bigl\{ \widehat{\mathcal{L}}_i(f) + \lambda\cdot{\widehat{\mathcal{L}}}^{(\gamma)}_{i,\text{\rm fair}}(f)\Bigr\} \\
    := &\min_{f\in\mathcal{F}}  \frac{1}{m}\sum_{i=1}^m  \min_{f_i \in\mathcal{F}}\Bigl\{\widehat{\mathcal{L}}_i(f) +\lambda {\widehat{\mathcal{L}}}_{\mathrm{fair},i}(f_i) + \frac{\lambda\gamma}{2} \bigl\Vert f_i - f\bigr\Vert^2\Bigr\}\nonumber 
\end{align}

As seen in this formulation, pFedFair builds upon the \emph{Moreau envelope approach}, applying it specifically to each client's fairness-aware loss function. Unlike pFedMe, which aims for full personalization, pFedFair ensures that the global model maintains a \emph{low total non-fairness-aware loss} while each personalized model improves the fairness of the global model. This approach is essential for achieving optimal performance in FL settings with client-level fairness constraints. Algorithm~\ref{algo:pFedFair} presents the pseudocode for pFedFair, which optimizes the weights $ w,w_i\in\mathbb{R}^d $ of the global classifier $ f_w $ and each personalized model $ f_{w_i} $.

\iffalse
\begin{figure}[H]
    \centering
    \includegraphics[width=0.99\linewidth]{KDE.png}
    \caption{KDE for Adult and COMPAS}
    \label{fig:KDE}
\end{figure}
\fi

\section{Numerical Results}

\subsection{Experimental Setup}

\begin{figure*}[htbp]
    \centering
    \subfloat[NPR of Fairness-aware FedAvg]{%
        \includegraphics[width=0.49\textwidth]{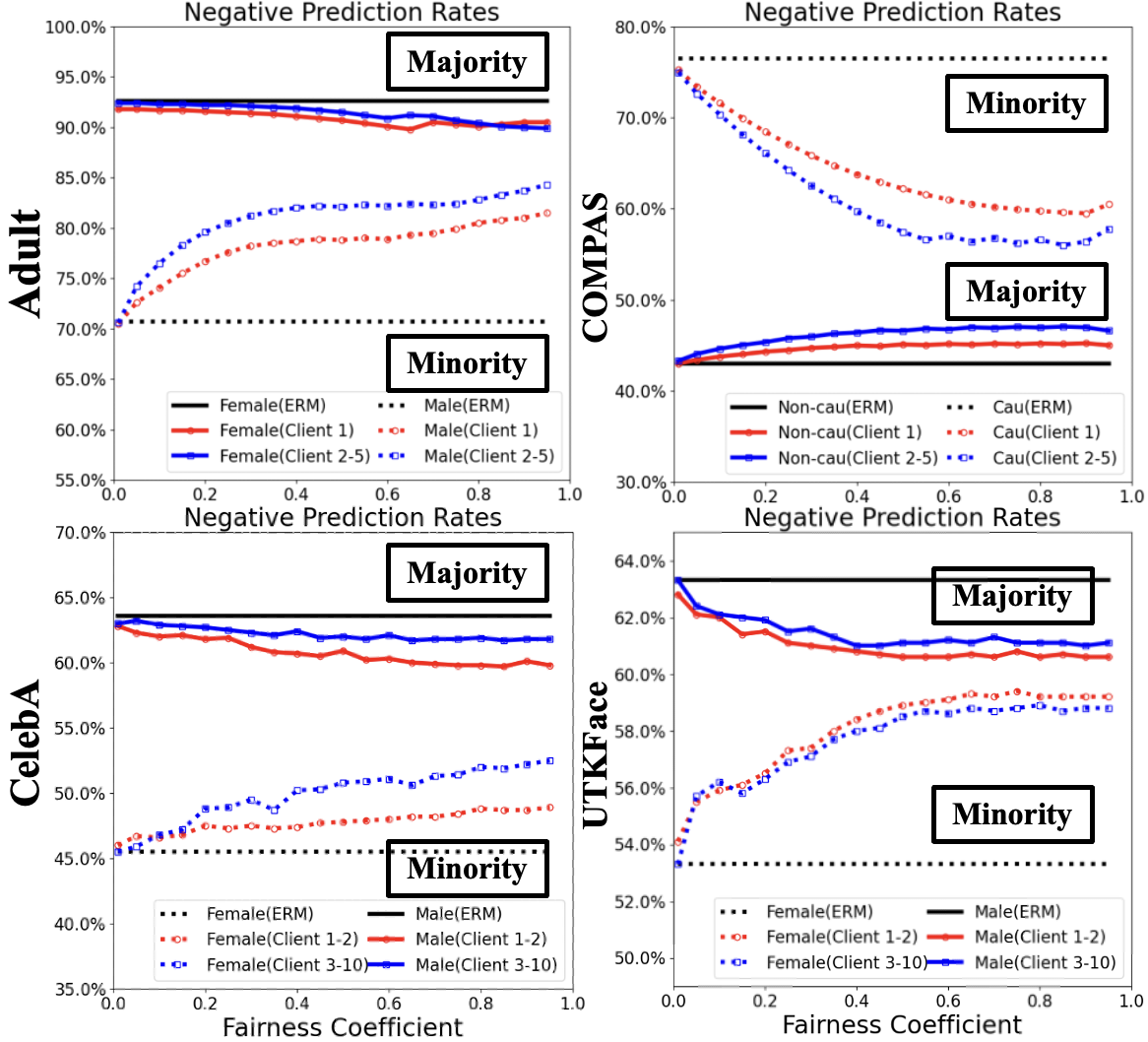}}%
    \subfloat[NPR of pFedFair]{%
        \includegraphics[width=0.49\textwidth]{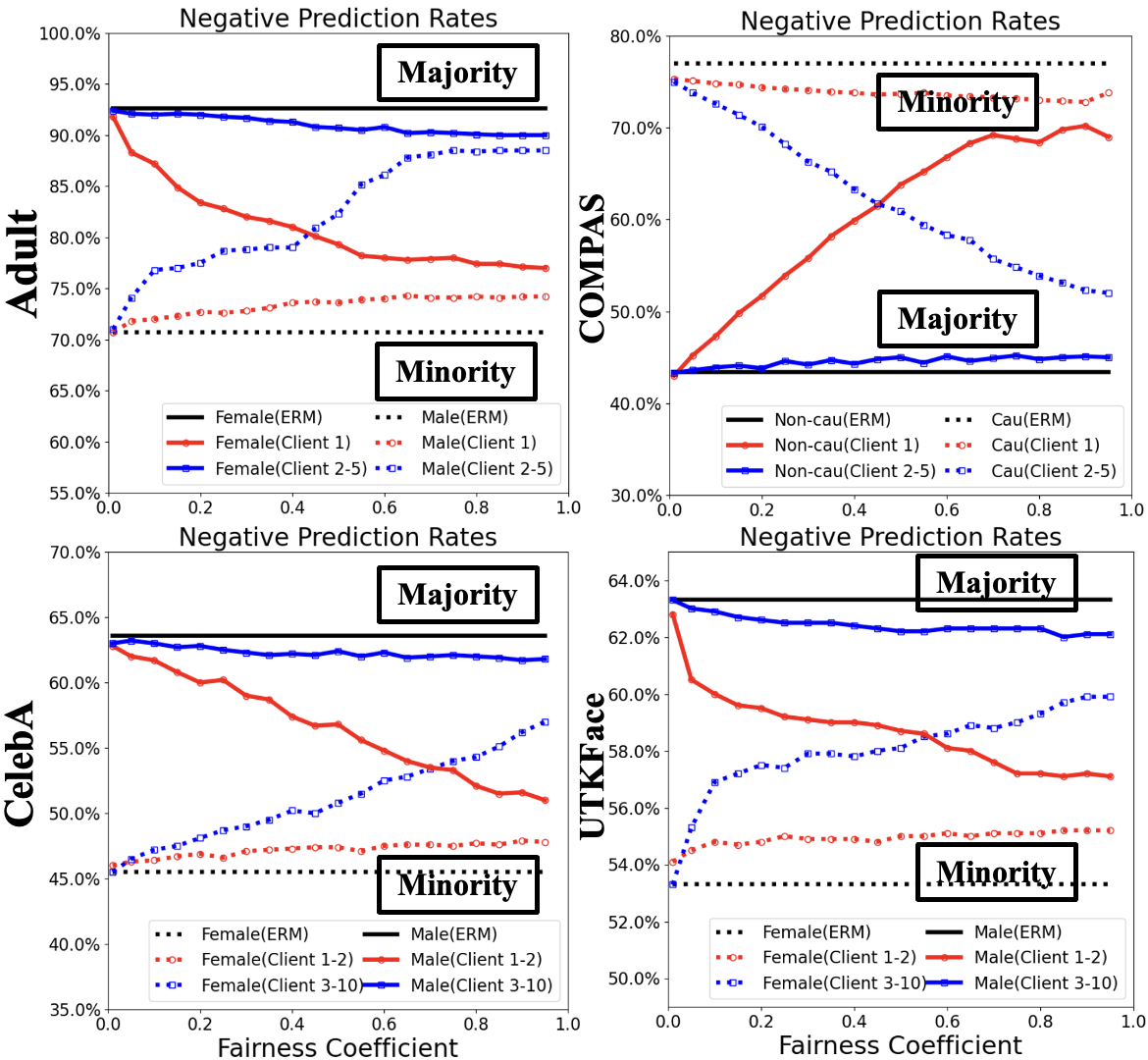}}%
    \caption{Experimental results comparing \textit{Negative Prediction Rates (NPR)} of different clients for Fairness-aware FedAvg (\textbf{Left}) and pFedFair (\textbf{Right}) algorithms when applied to the four benchmarks. The fairness coefficient $\eta=0$ means an ERM setting with no fairness constraint, while $\eta=0.9$ is the strongest fairness regularization coefficient over the range $[0,1)$.}
    \label{fig:induct_compas}
\end{figure*}

\textbf{Datasets.} We conduct experiments on four benchmark datasets widely adopted in fairness-aware learning, including both tabular and facial recognition datasets:

\begin{enumerate}[leftmargin=*]
\item \textbf{CelebA}~\cite{liu2018large}: A dataset of celebrity images with attribute annotations. "Smiling" is used as the binary label, and gender as the sensitive attribute. We used 12k training and 6k test samples.
\item \textbf{UTKFace}~\cite{zhang2017age}: A face dataset annotated with age, gender, and ethnicity. "Ethnicity" serves as the binary label, and gender as the sensitive attribute. The dataset was split into 8k training and 4k test samples.
\item \textbf{Adult}: A census dataset with 64 binary features and income labels ($>50K$ or not). Gender is the sensitive attribute. We used 12k/3k samples for training/testing.
\item \textbf{COMPAS}: A criminal recidivism dataset with 12 features, binary two-year recidivism labels, and race (Caucasian/Non-Caucasian) as the sensitive attribute, split into 3k/600 training/test samples.
\end{enumerate}

\noindent\textbf{Distributed Settings.} To evaluate fair FL under data heterogeneity, we partition each dataset into non-overlapping clients with distinct distributions of sensitive attributes. We divide clients as either \textit{"Underrepresented"} or \textit{"Overrepresented"} based on the alignment with the global distribution of sensitive attributes. Specifically,  a few clients are categorized as \textit{"Underrepresented"} as their majority sensitive attributes differ from the global distribution, while the rest are \textit{"Overrepresented"} as they align with it. To simulate realistic heterogeneity while maintaining average proportions, we introduce slight variations through Dirichlet distribution sampling; see Appendix.~\ref{appendix:setup} for more details.

\noindent\textbf{Baselines.} For fairness-aware algorithms, we adopt KDE~\cite{cho2020bfair}, a state-of-the-art method for fair learning in centralized settings, as the foundational approach. In fairness-aware FL settings, we evaluate: (1) localized fairness-aware training: KDE (local), (2) fairness-aware model aggregation: FedAvg~\cite{mcmahan2017communication}+KDE, and (3) fairness-aware personalized FL: pFedMe~\cite{t2020personalized}+KDE. For image datasets, we use DINOv2~\cite{oquab2023dinov2} and CLIP~\cite{radford2021learning} as embedding models and train a single linear layer for fair visual recognition tasks.

% For image experiments, we implement pFedFair using embeddings from foundation models, specifically DINOv2~\cite{oquab2023dinov2} and CLIP~\cite{radford2021learning}, coupled with a single-layer linear classifier. In centralized training scenarios, we establish multiple baseline models with fairness regularization: ResNet-18~\cite{he2016deep}, ViT-B/16~\cite{dosovitskiy2020image}, self-implemented CNN, and MLP. We also compare with several state-of-the-art fair face recognition methods, including IRM~\cite{rosenfeld2020risks}, RES~\cite{romano2020achieving}, DRO~\cite{levy2020large}, and DiGA~\cite{zhang2024distributionally}. Additionally, we implement KDE(local), FedAvg+KDE, and pFedMe+KDE in federated learning settings; See the Appendix for details.

% For tabular datasets, we implement KDE(local), FedAvg+KDE, pFedMe+KDE, FA~\cite{du2021fairness}, MMPF~\cite{martinez2020minimax}, FCFL~\cite{cui2021addressing}, and our proposed pFedFair for comprehensive comparison. These federated fairness baselines aim to improve the client-level fairness-accuracy trade-off.

\noindent\textbf{Evaluation Metrics.} We evaluate our model using four criteria: (1) \textbf{Utility}: \emph{Accuracy} (Acc) and \emph{Test Error} (1-Acc) measure the model's predictive performance. (2) \textbf{Fairness}: We use the \emph{Difference of Demographic Parity (DDP)}~\cite{dwork2012fairness}, which quantifies disparities in positive predictions across demographic groups. (3) \textbf{Consistency}: Following~\cite{li2019fair, cui2021addressing}, we report the worst-case utility and fairness metrics across all clients to evaluate client-level trade-offs in federated settings. (4) \textbf{Global Model Biases}: \emph{Negative Prediction Rates} (NPR) conditioned on sensitive attributes, defined as $\mathrm{NPR}(s) := P\bigl(\hat{Y}=0\mid S=s\bigr)$, is used to measure systematic disparities in negative predictions across subgroups.
\subsection{Biases of Global Classifier in Heterogeneous FL}

We first present a comparative analysis of model aggregation w/wo personalization across diverse datasets to show the impact of biases on fair federated learning~\cite{lei2024inductive}. NPR is used as the metric. As shown in Fig.~\ref{fig:induct_compas}, there is a key issue with global model predictions: without personalization, increasing fairness regularization drives the NPR to align with the majority sensitive attribute ($S=s_{\mathrm{max}}$) across the dataset. This alignment introduces bias against underrepresented clients, especially when their local sensitive attribute distribution differs from the global majority. NPR analysis shows that such biases in global fair classifiers increase misclassification rates for minority clients and lead to suboptimal fairness-accuracy trade-offs under heterogeneous demographics. In contrast, our proposed pFedFair framework mitigates this issue by incorporating fairness-aware personalization, adaptively optimizing local models to achieve better client-level accuracy-fairness trade-offs.

\begin{figure*}[htbp]
    \centering
    \includegraphics[width=0.98\textwidth]{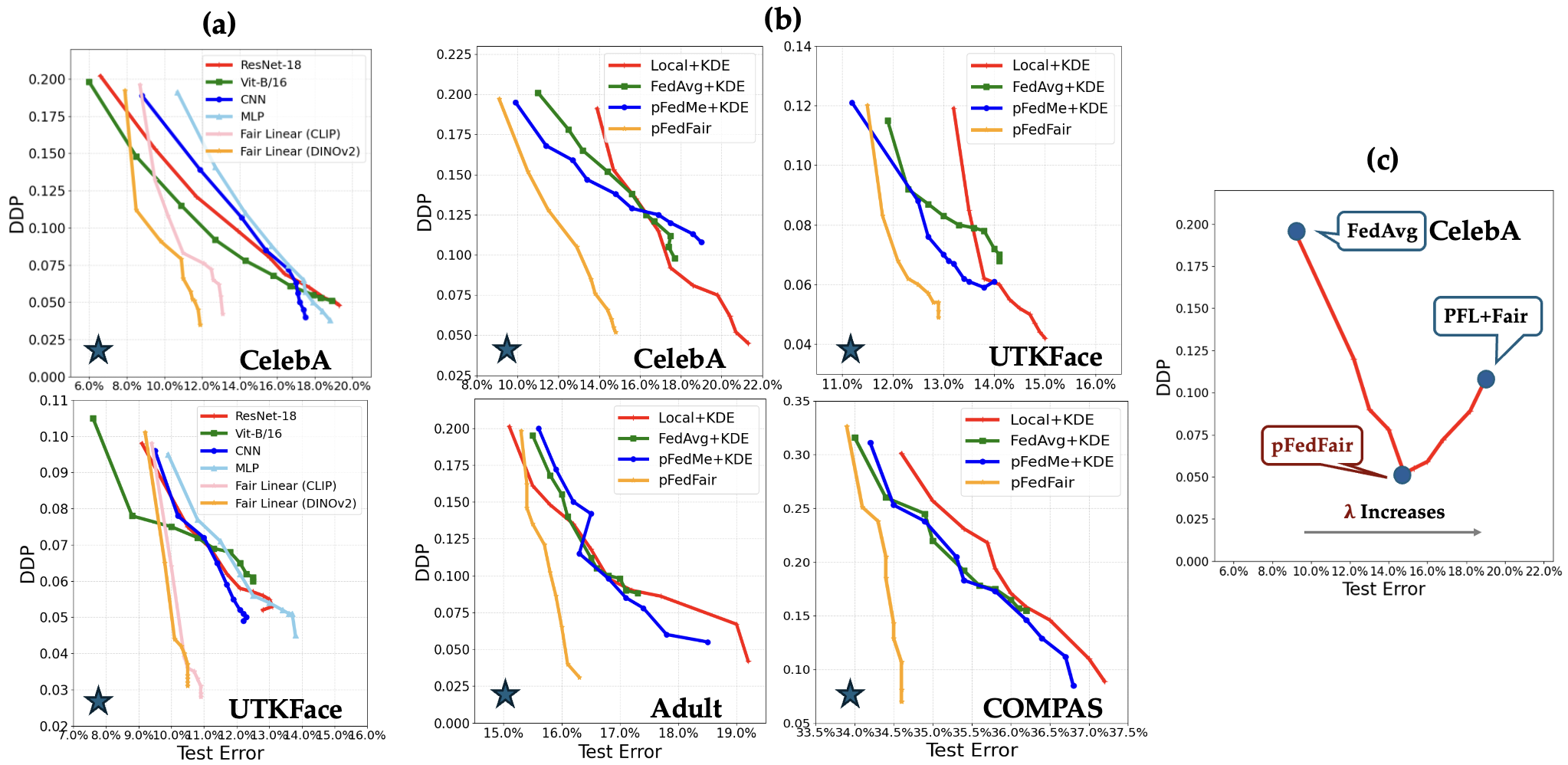}
    \caption{\textbf{Experimental Results.} (a) Test Error ($\downarrow$) vs. DDP ($\downarrow$) trade-off in centralized settings, compared to other fairness-aware visual recognition frameworks. (b) Test Error ($\downarrow$) vs. DDP ($\downarrow$) trade-off in federated learning settings, compared to other fairness-aware FL frameworks. (c) Effect of parameter $\lambda$ in balancing global model updates and local fairness-aware personalization for optimal trade-offs.}
    \label{fig: experiment}
\end{figure*}

% \subsection{Fair Visual Recognition via Embeddings}
\subsection{Fairness-aware Learning in Centralized Cases}

% Although state-of-the-art fairness-aware algorithms have achieved notable success on tabular benchmarks such as Adult and COMPAS, their effectiveness in high-dimensional vision tasks remains largely underexplored. 

% \begin{figure}[htbp]
%     \centering
%     \subfloat[Centralized Trade-off on CelebA]{%
%     \includegraphics[width=0.24\textwidth]{central_celeba.png}}%
%     \subfloat[Centralized Trade-off on UTKFace]{%
%     \includegraphics[width=0.24\textwidth]{central_utk.png}}%
%     \caption{Comparison of different fair image recognition models on CelebA and UTKFace datasets. We show the trade-off between Test Error (↓) and DDP (↓) when incorporating fairness algorithms. The lower left position indicates a superior trade-off.}
%     \label{fig:central_celeba}
% \end{figure}

Fairness-aware algorithms utilize various dependence measures for batch-wise fairness regularization, such as Mutual Information~\cite{cho2020afair}, Maximal Correlation~\cite{mary2019fairness}, and Exponential Rényi Mutual Information~\cite{lowy2022stochastic}. However, in visual recognition tasks, these regularizers often produce biased estimations due to batch size constraints. To address this limitation, we leverage pre-trained DINOv2~\cite{oquab2023dinov2} and CLIP~\cite{radford2021learning} as feature extractors, combined with a linear classifier trained using KDE-based fairness estimation~\cite{cho2020bfair}. This approach supports larger batch sizes, resulting in more accurate fairness estimations for visual recognition tasks. In this section, we conduct experiments on two large-scale image datasets: CelebA~\cite{liu2018large} and UTKFace~\cite{zhang2017age}. In Fig.~\ref{fig: experiment}(a), we present a numerical evaluation of our proposed Fair Linear Classifier (Fair Linear) using embeddings from DINOv2 and CLIP, compared to common image classification baselines: ResNet-18~\cite{he2016deep}, ViT-B/16~\cite{radford2021learning}, CNN, and MLP, all incorporating KDE for fairness-aware learning. Our results demonstrate that the linear classifier trained on these embeddings consistently achieves superior fairness-accuracy trade-offs on CelebA and UTKFace. This improvement is consistent across fairness coefficient ($\eta$), ranging from 0 to 0.9, while significantly reducing computational overhead.

\subsection{Fairness-aware Learning in Heterogeneous FL} 

% \begin{figure}[htbp]
%     \centering
%     \includegraphics[width=0.3\textwidth]{lambda.png}
%     \caption{Worst-case Test Error ($\downarrow$)/DDP ($\downarrow$) on CelebA dataset with variant hyperparameter $\lambda$. It controls the trade-off between global model utility and local fairness-aware personalization.}
%     \label{fig:lambda_celeba}
% \end{figure}

\textbf{Classification on image datasets.} We then show results of fair image classification in FL settings. Few prior works in fair FL address visual recognition on image datasets, as fairness-aware algorithms often fail to converge with complex architectures (e.g., ResNet, ViT) under heterogeneous and limited data availability in FL settings. Inspired by embedding-based fair image classification methods, we extend this approach to our proposed pFedFair. We evaluate pFedFair on heterogeneously distributed CelebA and UTKFace datasets, simulating biased distributions where Clients 1-2 ("Underrepresented") and Clients 3-10 ("Overrepresented") have inverse proportions of sensitive groups. Fig.~\ref{fig: experiment}(b) compares pFedFair against three baselines: Localized fairness-aware learning (KDE), fairness-aware global aggregation (FedAvg+KDE), and fairness-aware personalized FL (pFedMe+KDE), focusing on the client-level worst-case Test Error/DDP trade-off. All methods utilize DINOv2 embeddings with a linear classifier and KDE, with $\lambda=0.4$ balancing global and local updates. Our results show that pFedFair consistently outperforms all baselines in achieving optimal client-level Test Error/DDP trade-offs on CelebA, UTKFace, and the tabular datasets Adult and COMPAS. Besides, in Fig.~\ref{fig: experiment}(c), we show how the parameter $\lambda$ balances the global model updates and local fairness-aware personalization to explore an optimal trade-off. By leveraging pre-trained embeddings, pFedFair reduces computational overhead while ensuring superior fairness-accuracy trade-offs, establishing it as an efficient solution for fair visual recognition in heterogeneous FL settings.

% \begin{figure}[htbp]
%     \centering
%     \subfloat[Trade-off on CelebA]{%
%     \includegraphics[width=0.24\textwidth]{worst_celeba.png}}% 
%     \subfloat[Trade-off on UTKFace]{%
%     \includegraphics[width=0.24\textwidth]{worst_utk.png}}%
%     \\
%     \subfloat[Trade-off on Adult]{%
%     \includegraphics[width=0.24\textwidth]{worst_adult.png}}%
%     \subfloat[Trade-off on COMPAS]{%
%     \includegraphics[width=0.24\textwidth]{worst_compas.png}}%
%     \caption{Analysis of the average and the worst-performing client's trade-off between Test Error (↓) and DDP (↓) at the client level, evaluated on CelebA and FairFace datasets. The lower left position indicates a superior trade-off.}
%     \label{fig:fl_celeba}
% \end{figure}

\begin{table*}[t]\LARGE
    \centering
    \renewcommand{\arraystretch}{1.25}
    \resizebox{0.99\textwidth}{!}{
    \tabcolsep=0.5cm
    \begin{tabular}{clcccccccccc}
    \toprule
    & & \multicolumn{2}{c}{\textbf{Client 1 (Underrep.)}} & \multicolumn{2}{c}{\textbf{Client 2 (Overrep.)}} & \multicolumn{2}{c}{\textbf{Client 3 (Overrep.)}} & \multicolumn{2}{c}{\textbf{Client 4 (Overrep.)}} & \multicolumn{2}{c}{\textbf{Client 5 (Overrep.)}}\\
    \cmidrule(l){3-4}
    \cmidrule(l){5-6}
    \cmidrule(l){7-8}
    \cmidrule(l){9-10}
    \cmidrule(l){11-12}
     & & \textbf{Acc($\uparrow$)} & \textbf{DDP($\downarrow$)} & \textbf{Acc($\uparrow$)} & \textbf{DDP($\downarrow$)} & \textbf{Acc($\uparrow$)} & \textbf{DDP($\downarrow$)}
     & \textbf{Acc($\uparrow$)} & \textbf{DDP($\downarrow$)} & \textbf{Acc($\uparrow$)} & \textbf{DDP($\downarrow$)}\\

    \cmidrule[1.5pt]{2-12}

    \parbox[t]{2mm}{\multirow{7}{*}{\rotatebox[origin=c]{90}{\text{\LARGE{\textbf{Adult}}}}}}
    &ERM(local) & 84.5\% & 0.198 & 87.0\% & 0.252 & 86.9\% & 0.262 & 86.5\% & 0.328 & 87.0\% & 0.286\\
    &FedAvg & 85.1\% & 0.188 & 87.1\% & 0.310 & 87.1\% & 0.252 & 87.6\% & 0.256 & 87.1\% & 0.244\\
    &pFedMe & 85.5\% & 0.231 & 87.4\% & 0.299 & 87.0\% & 0.278 & 87.9\% & 0.248 & 87.7\% & 0.301\\
    \cmidrule[0.4pt]{2-12}
    &KDE(local) &81.9\% & 0.037 & 85.2\% & 0.038 & 85.0\% & 0.020 & 85.7\% & 0.032 & 85.5\% & 0.031 \\
    &FedAvg+KDE & 81.9\% & 0.104 & 85.5\% & 0.073 & \textbf{86.6\%} & 0.058 & 86.7\% & 0.063 & 86.2\% & 0.060\\
    &pFedMe+KDE &81.4\% & 0.070 & 84.1\% & 0.038 &85.0\% & 0.039 &85.1\% & 0.023 & 85.0\% & 0.035\\
    &\textbf{pFedFair} & \textbf{84.7\%} & \textbf{0.031} & \textbf{86.5\%} & \textbf{0.035} & 86.5\% & \textbf{0.012} & \textbf{87.1\%} & \textbf{0.014} & \textbf{86.6\%} & \textbf{0.030}\\
    
    \cmidrule[1.5pt]{2-12}

    \parbox[t]{2mm}{\multirow{7}{*}{\rotatebox[origin=c]{90}{\text{\LARGE{\textbf{COMPAS}}}}}}
    &ERM(local) &64.5\% & 0.251 & 66.8\% & 0.251 & 66.1\% & 0.275 & 66.0\% & 0.340 & 66.4\% & 0.278\\
    &FedAvg &65.8\% & 0.291 & 67.7\% & 0.301 & 67.4\% & 0.308 & 67.7\% & 0.275 & 67.8\% & 0.301\\
    &pFedMe & 66.1\% & 0.311 & 67.5\% & 0.279 & 67.9\% & 0.282 & 68.0\% & 0.293 & 68.2\% & 0.307\\
    \cmidrule[0.4pt]{2-12}
    &KDE(local) &63.0\% & \textbf{0.045} & 65.6\% & 0.089 & 64.7\% & 0.088 & 64.7\% & 0.075 &  64.8\% & 0.076\\
    &FedAvg+KDE &63.8\% & 0.155 &66.4\% & 0.145 & 65.6\% & 0.153 &65.5\% & 0.108 & 66.1\% & 0.111\\
    &pFedMe+KDE &63.2\% & 0.088 & 65.0\% & 0.077 & 64.2\% & 0.090 & 64.1\% &0.084 & 64.9\% & 0.082\\
    &\textbf{pFedFair} & \textbf{65.4\%} & 0.051 & \textbf{67.2\%} & \textbf{0.065}& \textbf{66.8\%} & \textbf{0.068} & \textbf{66.9\%} & \textbf{0.060} & \textbf{67.1\%} & \textbf{0.059}\\

    \cmidrule[1.5pt]{2-12}
    
    \end{tabular}}
    \caption{Client-level experimental results on Adult and COMPAS demonstrate the efficacy of pFedFair. The fairness constraint coefficient is set to be $\eta=0.9$. \textit{Overrep.} and \textit{Underrep.} represent the overrepresented clients and underrepresented clients respectively.}
    \label{table_client}
\end{table*}

\begin{figure*}[htbp]
    \centering
    \subfloat[FedAvg+KDE]{%
        \includegraphics[width=0.16\textwidth]{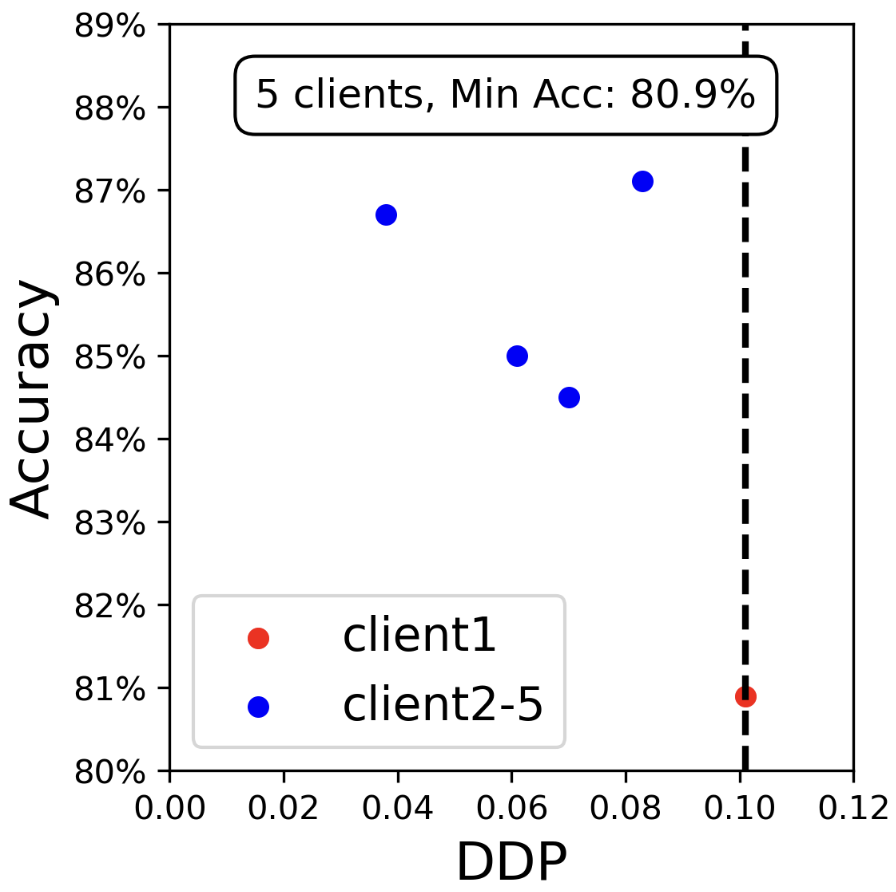}}%
    \subfloat[pFedMe+KDE]{%
        \includegraphics[width=0.16\textwidth]{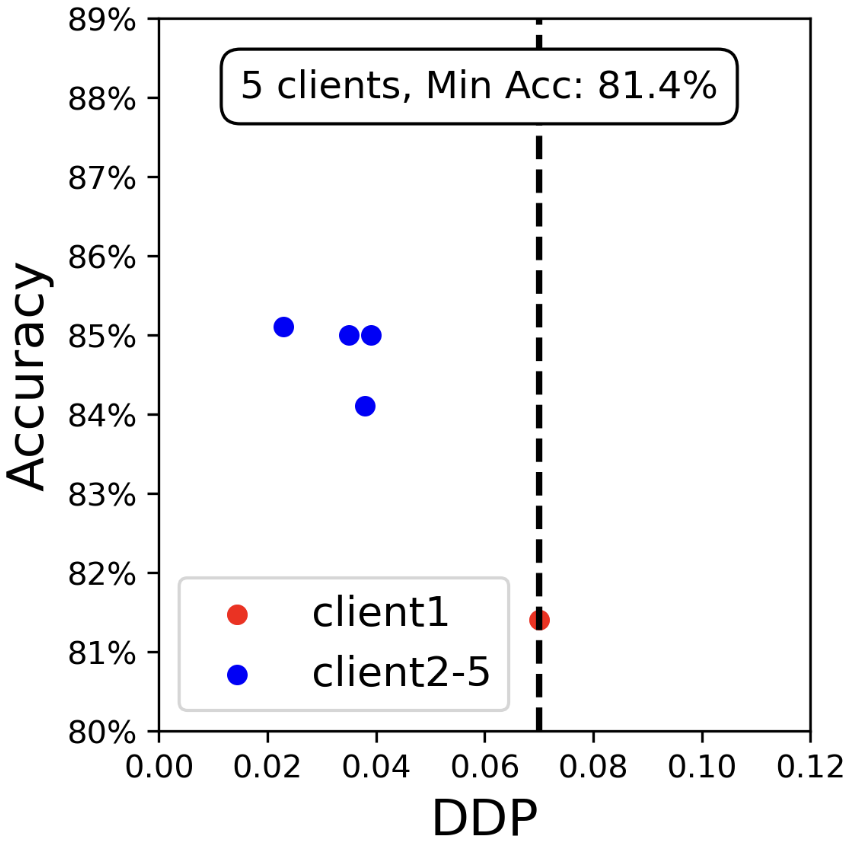}}%
    \subfloat[FA]{%
        \includegraphics[width=0.16\textwidth]{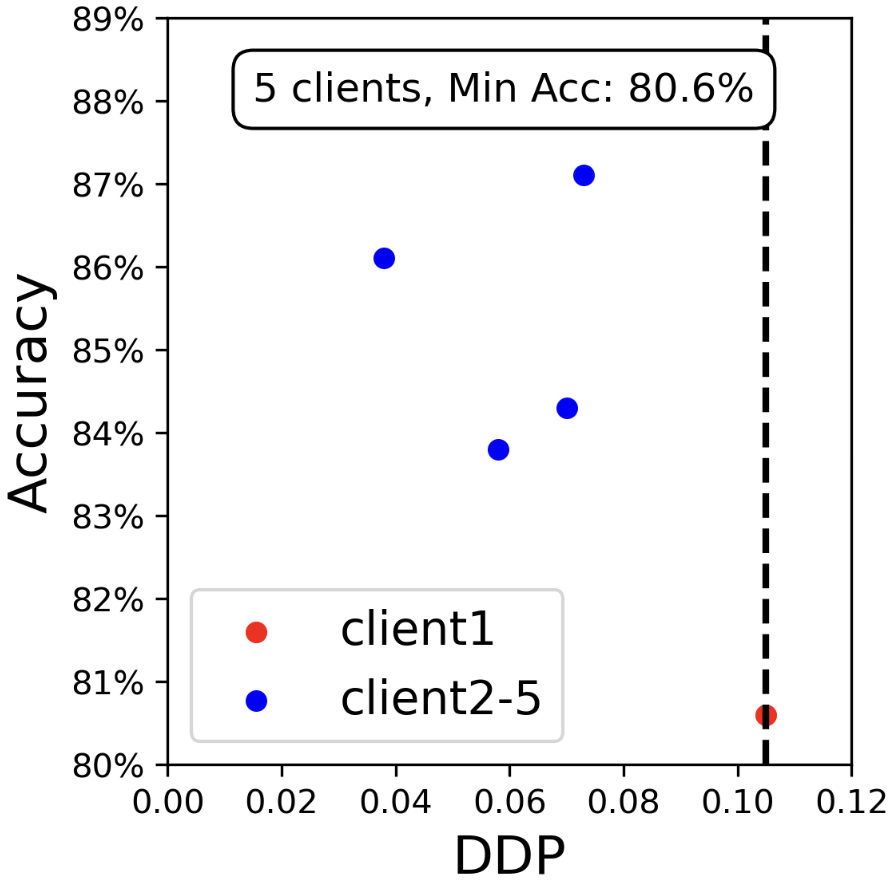}}%
    \subfloat[MMPF]{%
        \includegraphics[width=0.16\textwidth]{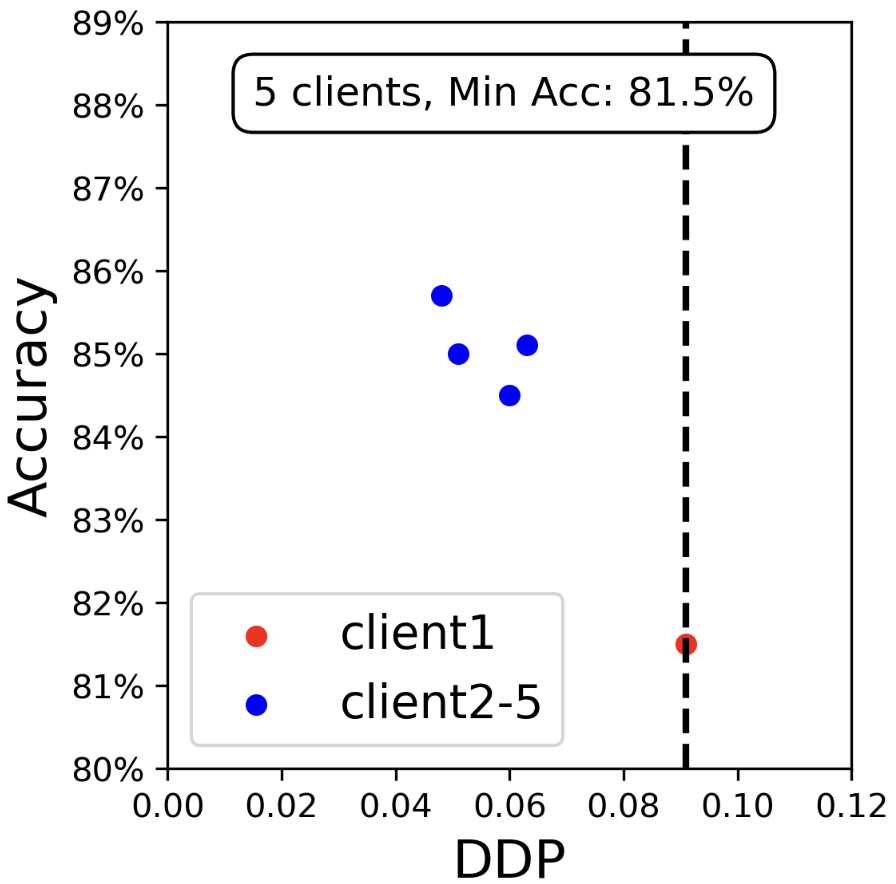}}%
    \subfloat[FCFL]{%
        \includegraphics[width=0.16\textwidth]{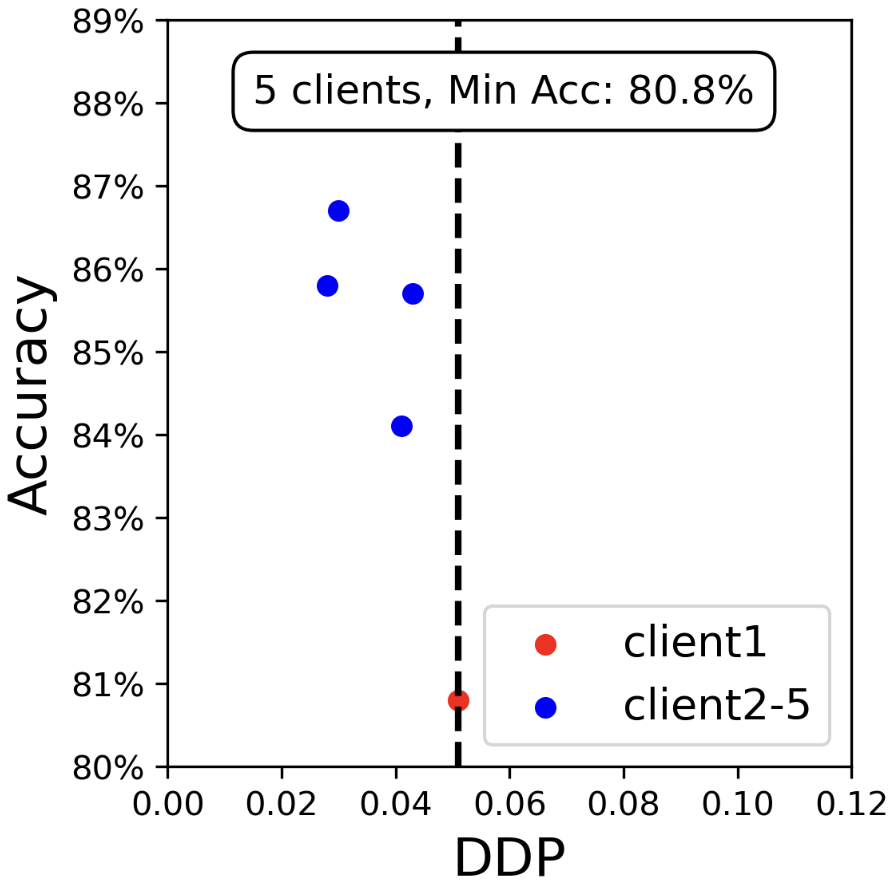}}%
    \subfloat[\textbf{pFedFair}]{%
        \includegraphics[width=0.16\textwidth]{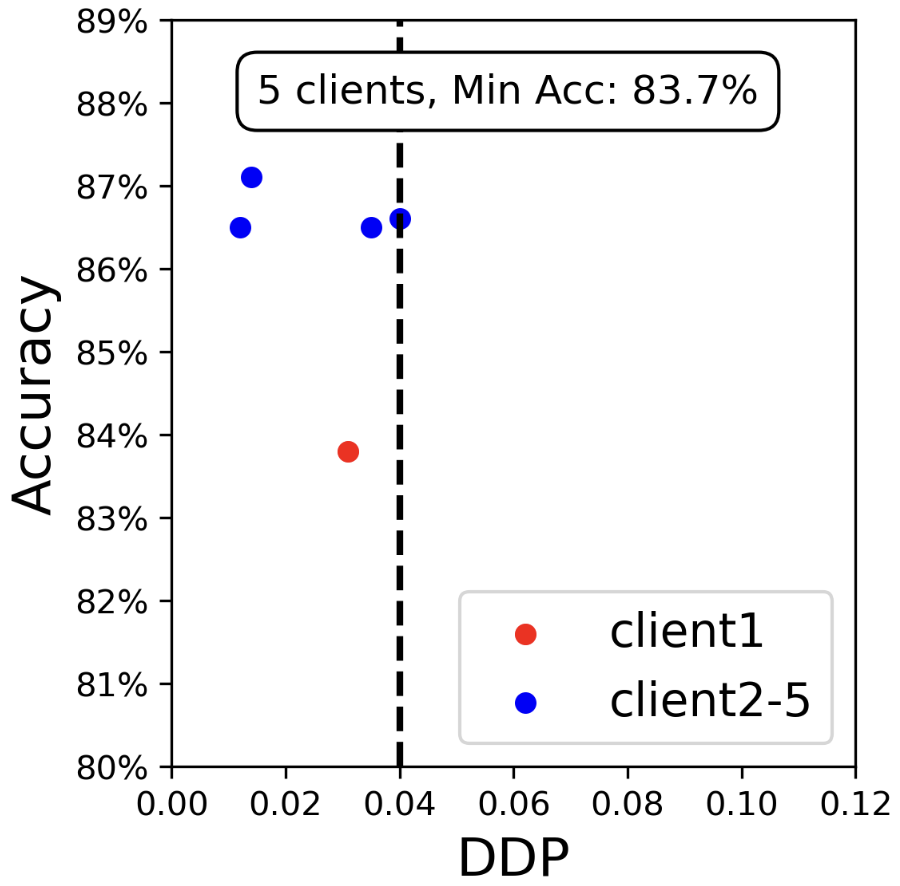}}%
    
    \centering
    \subfloat[FedAvg+KDE]{%
        \includegraphics[width=0.16\textwidth]{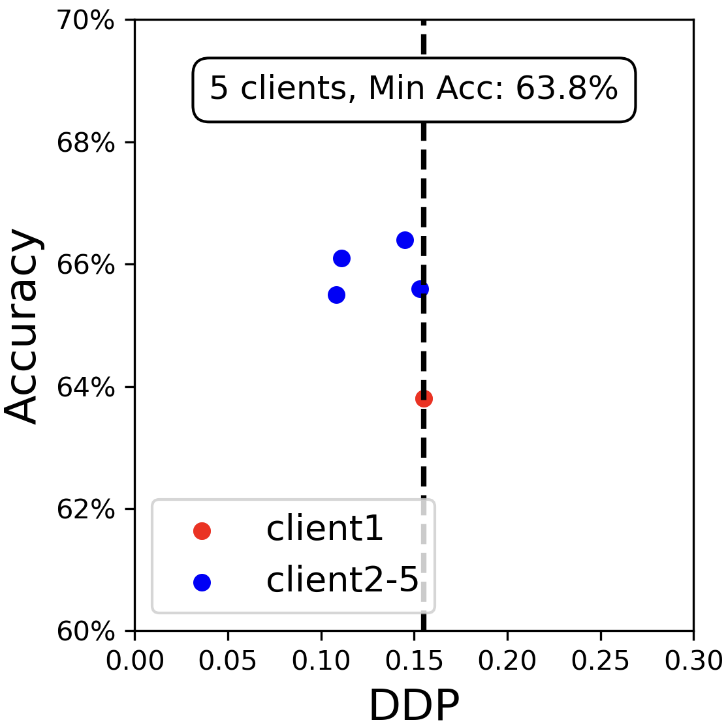}}%
    \subfloat[pFedMe+KDE]{%
        \includegraphics[width=0.16\textwidth]{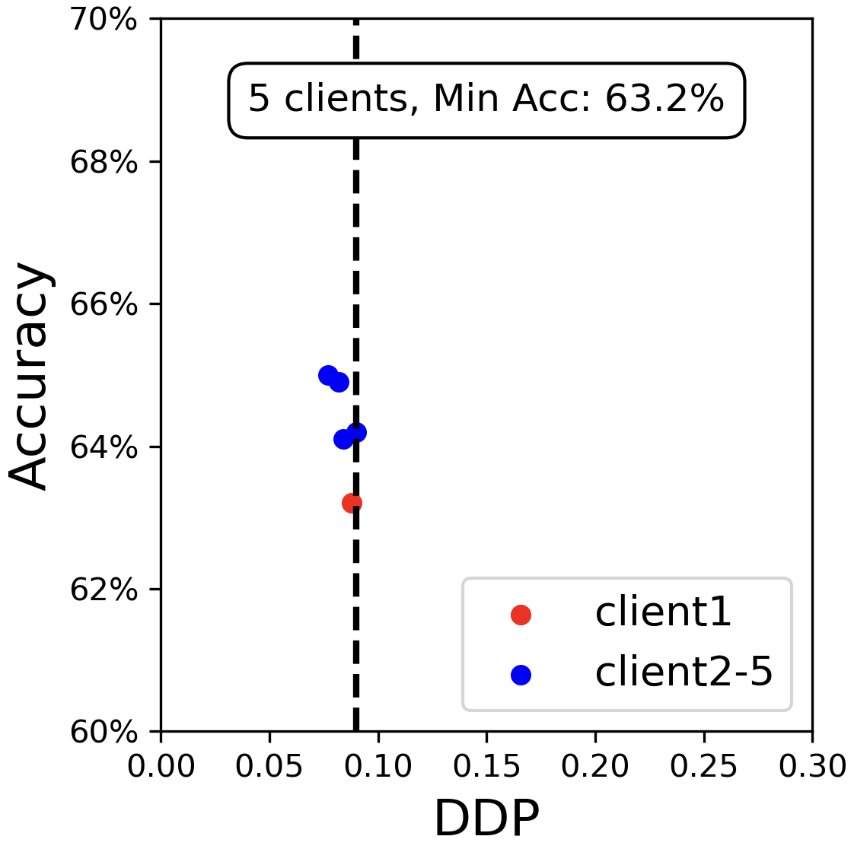}}%
    \subfloat[FA]{%
        \includegraphics[width=0.16\textwidth]{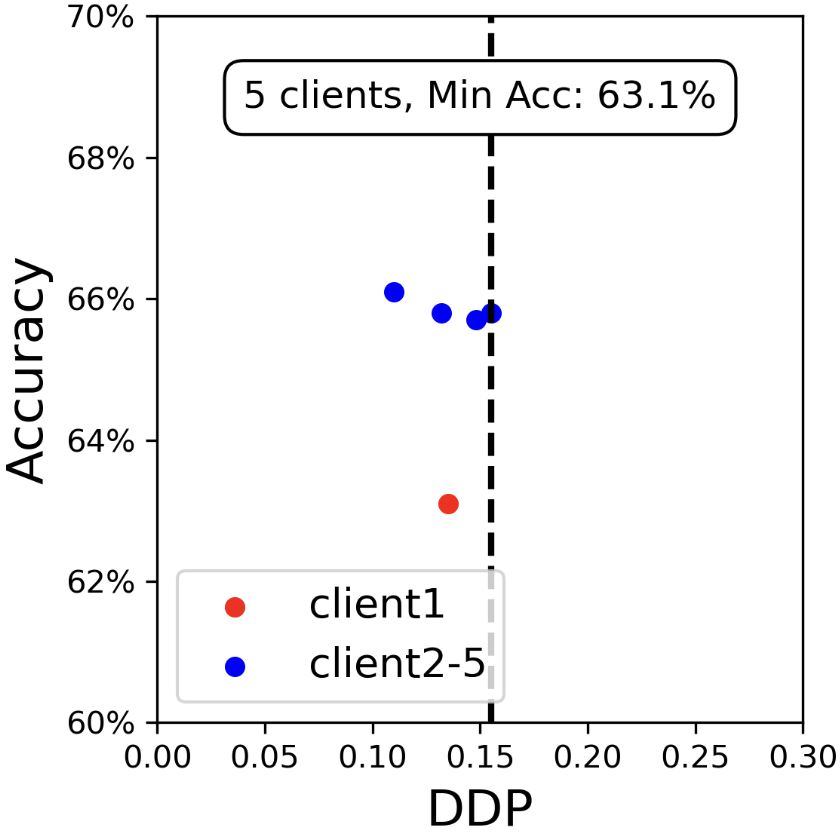}}%
    \subfloat[MMPF]{%
        \includegraphics[width=0.16\textwidth]{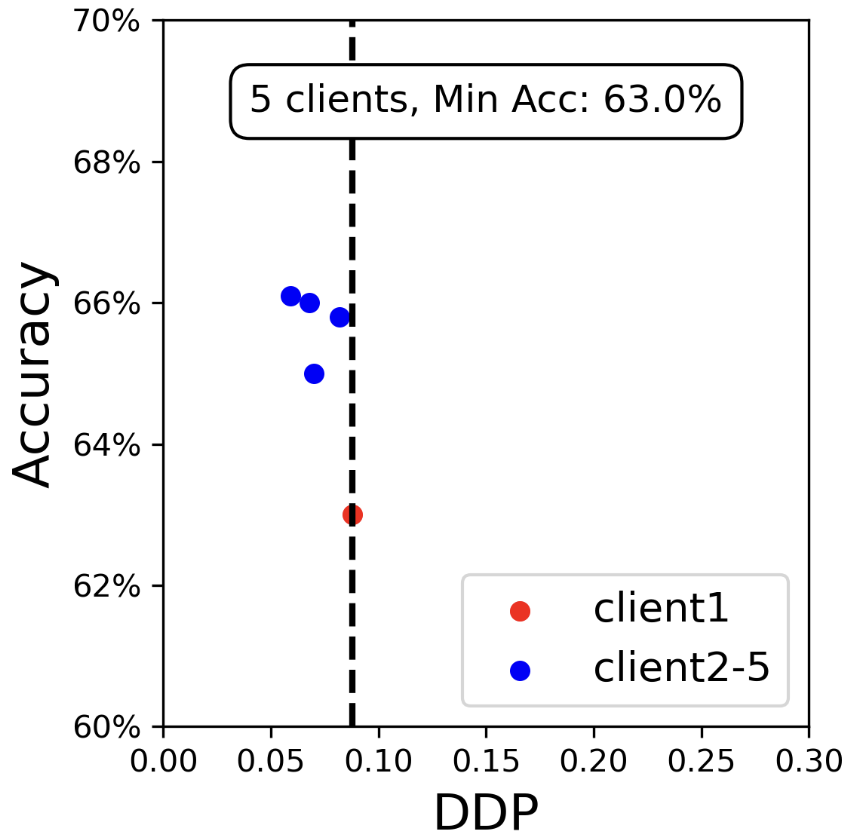}}%
    \subfloat[FCFL]{%
        \includegraphics[width=0.16\textwidth]{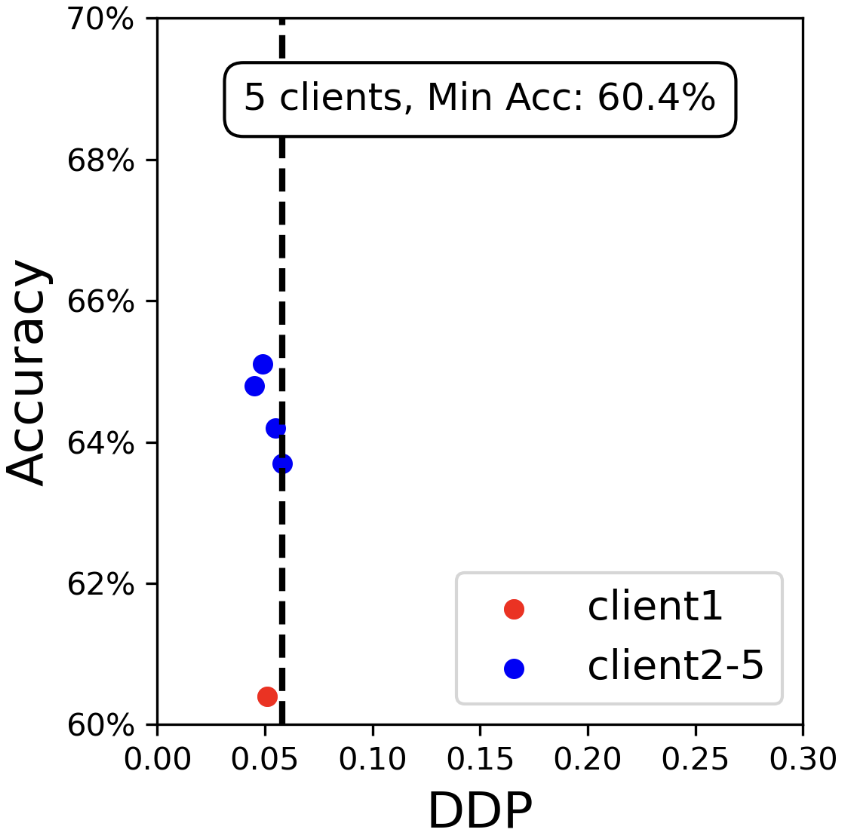}}%
    \subfloat[\textbf{pFedFair}]{%
        \includegraphics[width=0.16\textwidth]{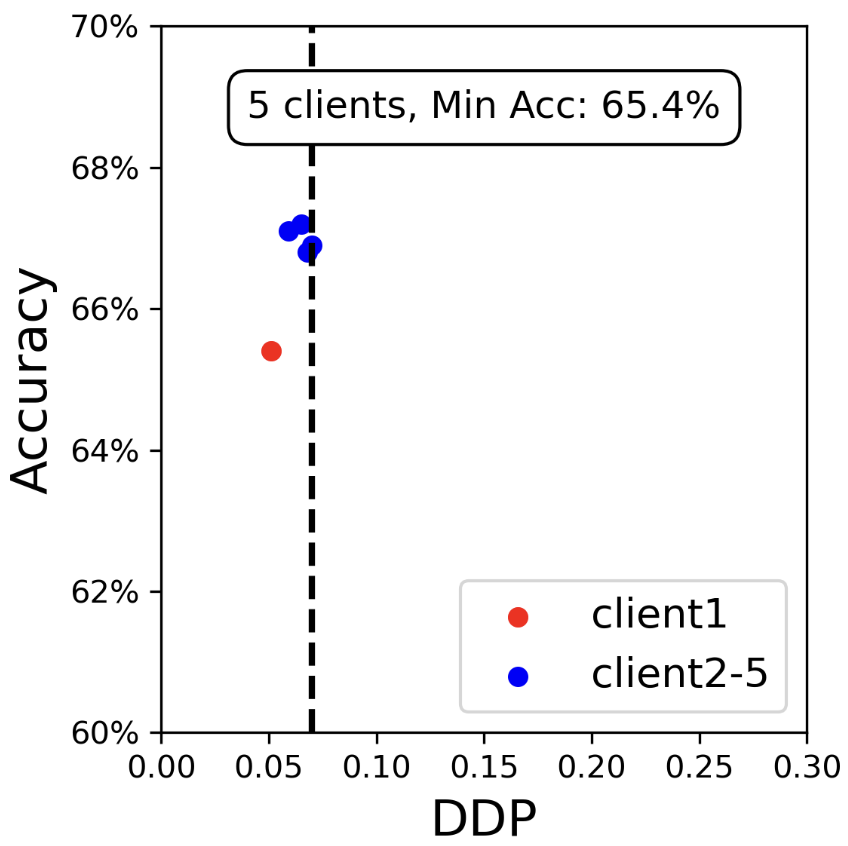}}%
    \caption{Experimental results of different baseline methods on Adult (\textbf{top}) and COMPAS (\textbf{bottom}). Each data point represents the client-level performance on Accuracy and DDP.}
    \label{fig:compas}
\end{figure*}

\noindent\textbf{Classification on tabular datasets.} 
We evaluate pFedFair on two tabular datasets, Adult and COMPAS, under heterogeneous FL settings. Table~\ref{table_client} presents client-specific accuracy and DDP. Under ERM, both localized training and FedAvg provide marginal accuracy gains while maintaining consistent DDP across clients. However, for underrepresented Client 1, whose sensitive attribute distribution deviates from the global distribution, global fair algorithms align predictions with overrepresented clients, causing significant performance drops compared to its locally-trained fair model. Our proposed pFedFair framework improves worst-case accuracy-DDP trade-offs across clients, and outperforms locally-trained KDE, FedAvg+KDE, and pFedMe+KDE with preserved privacy guarantees. We also compare pFedFair with other baselines targeting client-level group fairness-accuracy trade-offs, including FA \cite{du2021fairness}, MMPF \cite{martinez2020minimax}, and FCFL \cite{cui2021addressing}. Fig.~\ref{fig:compas} highlights pFedFair's superiority in client-level performances, achieving optimal accuracy-DDP trade-offs across all baselines. These findings validate pFedFair's ability to deliver consistent performance for all clients while ensuring optimal fairness-accuracy trade-offs in heterogeneous cases.

\section{Conclusion}

In this work, we introduced pFedFair, a personalized federated learning framework designed to enforce client-level group fairness in heterogeneous FL settings. We demonstrated that traditional FL approaches, which optimize a global model, could fail to balance accuracy and group fairness when clients have diverse distributions of sensitive attributes. By allowing clients to independently impose fairness constraints while still benefiting from shared knowledge, pFedFair achieved an improved accuracy-fairness trade-off in our experiments. Furthermore, by leveraging image embedding models, we extended the applicability of our method to computer vision tasks, indicating improvements over conventional group-fairness-aware learning strategies applied to computer vision data. Future work includes extending pFedFair to more complex vision tasks, such as object detection and generative models, and exploring adaptive fairness constraints tailored to multimodal vision-language models.

\bibliographystyle{unsrt}
\bibliography{main}

\begin{appendices}

\section{Appendix}
\subsection{Proof of Proposition \ref{Thm: 1}}
Part (a) of Proposition \ref{Thm: 1}  follows directly from the definition of Bayes classifier. Note that the Bayes classifier, i.e. the decision rule that minimizes the expected 0/1-loss (the misclassification probability), is given by $
    g(x,s) = \arg\!\max_{y} P^{(i)}(Y = y | X = x, S = s)$.

\noindent \textbf{Proof of Proposition \ref{Thm: 1}.(b)} For each Client $i$, we can write the following for the optimal fairness-aware classification rule $f_i(X)$, that is statistically independent of $Y$,
\begin{align*}
    P^{(i)}(f_i(X)\neq Y) \, =\, P^{(i)}(S=s^{(i)}_{\min})P^{(i)}(f_i(X)\neq Y|S=s^{(i)}_{\min}) + P^{(i)}(S=s^{(i)}_{\max})P^{(i)}(f_i(X)\neq Y|S=s^{(i)}_{\max})
\end{align*}
The total variation is the optimal transport cost corresponding to the 0/1 loss, i.e. it minimises the misclassification probability over all the joint couplings with the same marginals, and therefore we have $P^{(i)}(f_i(X)\neq Y|S=s) 
\ge \mathrm{TV}(P^{(i)}_{f_i(X)|S=s} , P^{(i)}_{Y|S=s})$, we obtain the following:
\begin{align*}
    P^{(i)}(f_i(X)\neq Y) \, &\ge \, P^{(i)}(S=s^{(i)}_{\min})\mathrm{TV}(P^{(i)}_{f_i(X)|S=s^{(i)}_{\min}}, P^{(i)}_{Y|S=s^{(i)}_{\min}}) + P^{(i)}(S=s^{(i)}_{\max})\mathrm{TV}(P^{(i)}_{f_i(X)|S=s^{(i)}_{\max}}, P^{(i)}_{Y|S=s^{(i)}_{\max}}) \\
    &\ge P^{(i)}(S=s^{(i)}_{\min})\mathrm{TV}(P^{(i)}_{Y|S=s^{(i)}_{\max}}, P^{(i)}_{Y|S=s^{(i)}_{\min}})
\end{align*}
where $\mathrm{TV}(\cdot , \cdot)$ denotes the total variation distance, and the inequalities hold with equality for the choice $P^{(i)}_{f_i(X)|S=s^{(i)}_{\min}} = P^{(i)}_{Y|S=s^{(i)}_{\max}}$ that minimizes the expected value of the total variation distance over the binary sensitive attribute $S\in\{0,1\}$.

On the other hand, in a federated learning setting with global function $f$ for all the clients that satisfy $f(X_i)\perp S_i$ at each client $i$, we will have the following:
\begin{align*}
    &\sum_{i=1}^m P^{(i)}(f(X)\neq Y)\\
    \, =\,  &\sum_{i=1}^mP^{(i)}(S=s^{(i)}_{\min})P^{(i)}(f(X)\neq Y|S=s^{(i)}_{\min}) + P^{(i)}(S=s^{(i)}_{\max})P^{(i)}(f(X)\neq Y|S=s^{(i)}_{\max}) \\
    \ge\,  &\sum_{i=1}^m P^{(i)}(S=s^{(i)}_{\min})\mathrm{TV}(P^{(i)}_{f(X)|S=s^{(i)}_{\min}} , P^{(i)}_{Y|S=s^{(i)}_{\min}}) + P^{(i)}(S=s^{(i)}_{\max})\mathrm{TV}(P^{(i)}_{f(X)|S=s^{(i)}_{\max}} , P^{(i)}_{Y|S=s^{(i)}_{\max}}) \\
    =\,  &\sum_{i=1}^m P^{(i)}(S=s^{(i)}_{\min})\mathrm{TV}(P_{f(X)} , P^{(i)}_{Y|S=s^{(i)}_{\min}}) + P^{(i)}(S=s^{(i)}_{\max})\mathrm{TV}(P_{f(X)} , P^{(i)}_{Y|S=s^{(i)}_{\max}}) \\
   \ge \,  &\sum_{i=1}^m P^{(i)}(S\neq s_{\max}) \mathrm{TV}(P^{(i)}_{Y|S=s_{\max}} , P^{(i)}_{Y|S\neq s_{\max}})  
\end{align*}
In the above, the inequalities will hold with equality and the optimality holds under the common marginal distribution $P_{f(X)} = P_{Y|S=s_{\max}}$. Therefore, since the probability of error matches the expected value of 0/1-loss, we will have
\begin{align*}
     &\sum_{i=1}^m \mathcal{L}_i(f) - \mathcal{L}_i(f^*_i)\\
    =\, & \sum_{i=1}^m P^{(i)}(f(X)\neq Y) - P^{(i)}(f_i^*(X)\neq Y) \\
    \ge\, & \sum_{i=1}^m P^{(i)}(S\neq s_{\max}) \mathrm{TV}(P^{(i)}_{Y|S=s_{\max}} , P^{(i)}_{Y|S\neq s_{\max}})  - P^{(i)}(S=s^{(i)}_{\min})\mathrm{TV}(P^{(i)}_{Y|S=s^{(i)}_{\max}}, P^{(i)}_{Y|S=s^{(i)}_{\min}}) \\
    =\, & \sum_{i=1}^m (2P^{(i)}(S= s^{(i)}_{\max})-1) \Bigl\vert \sum_{y\in\mathcal{Y}} P^{(i)}(Y=y|S= s_{\max})- P^{(i)}(Y=y|S= s^{(i)}_{\max})\Bigr\vert
\end{align*}
The final equality holds, because if $s_{\max} = s_{\max}^{(i)}$ at client $i$, both the terms on the two sides are zero, and if $s_{\max} = s_{\min}^{(i)}$, we can factor out the common total variation distance term and the remainder will be $P^{(i)}(S= s^{(i)}_{\max}) - P^{(i)}(S= s^{(i)}_{\min}) = 2P^{(i)}(S= s^{(i)}_{\max})-1$. Therefore, the proof is complete.

\subsection{Proof of Proposition 2}
To prove the proposition, we consider Client $i$'s fairness-aware problem formulation with the total variation loss function:
\begin{align*}
      &\min_{Q_{Y|X,S}}\quad\; \mathbb{E}_{P^{(i)}_{X,S}}\Bigl[\ell_{TV}(Q_{Y|X,S},P_{Y|X,S})\Bigr]\nonumber\\
      &\text{\rm subject to}\;\;\;\; \mathbb{E}_{s\sim P^{(i)}_S}[d(Q_{Y|S=s},Q_Y)]\le \epsilon. \nonumber  \end{align*}
Since the total variation is an optimal transport cost corresponding to the 0/1 loss, the above problem is equivalent to
\begin{align*}
      &\min_{Q_{Y|X,S}}\quad\; \mathbb{E}_{P^{(i)}_{X,S}}\Bigl[\min_{M_{\widehat{Y},Y}\in\Pi(Q_{Y|X,S},P_{Y|X,S})}\mathbb{E}_{M}\bigl[\ell_{0/1}(\widehat{Y},Y)\bigr]\Bigr]\nonumber\\
      &\text{\rm subject to}\;\;\;\; \mathbb{E}_{s\sim P^{(i)}_S}[d(Q_{Y|S=s},Q_Y)]\le \epsilon. \nonumber  \end{align*}
Note that in the optimization problem, we optimize the objective independently across different outcomes of $X=x,S=s$, and therefore, we can pull the minimization out as the minimization decouples across outcomes. This results in the following equivalent  formulation where :
\begin{align*}
      &\min_{Q_{Y|X,S}}\;\min_{M_{Y,\hat{Y}|X,S}\in \Pi(Q_{Y|X,S},P_{Y|X,S}) }\;\; \mathbb{E}_{P^{(i)}_{X,S}\times M_{Y,\hat{Y}|X,S} }\bigl[\ell_{0/1}(\widehat{Y},Y)\bigr]\nonumber\\
      &\text{\rm subject to}\;\;\;\; \mathbb{E}_{s\sim P^{(i)}_S}[d(Q_{Y|S=s},Q_Y)]\le \epsilon.  \end{align*}
Merging the two minimization subproblems into a single minimization task, we obtain the fairness-aware-constrained optimal transport formulation stated in the theorem:
    \begin{align*}
      &\min_{Q_{\widehat{Y},Y|X,S}}\quad\; \mathbb{E}_{P^{(i)}_{X,S}\times Q_{\widehat{Y},Y|X,S}}\Bigl[\ell_{0/1}(\widehat{Y},Y)\Bigr]\nonumber\\
      &\text{\rm subject to}\;\;\;\; Q_{Y|X,S}=P_{Y|X,S}\\
      &\qquad\qquad\;\;\;\: \mathbb{E}_{s\sim P^{(i)}_S}[d(Q_{\widehat{Y}|S=s},Q_{\widehat{Y}})]\le \epsilon. \nonumber  \end{align*}

\subsection{Details of Experimental Setup}
\label{appendix:setup}
\begin{itemize}
    \item \textbf{Datasets.} We conduct  experiments on four benchmark datasets widely adopted in fair machine learning research, including both tabular and facial recognition datasets:\\

\begin{enumerate}[leftmargin=*]
\item \textbf{CelebA}~\cite{liu2018large}, containing the pictures of celebrities with 40 attribute annotations, where we considered "Smiling" as a binary label, and the sensitive attribute is the binary variable on gender. In the experiments, we used 12,000 training samples and 6,000 test samples.\\
\item \textbf{UTKFace}~\cite{zhang2017age}, is a large-scale face dataset with annotations of age, gender, and ethnicity. We chose "Ethnicity" as a binary label, and the sensitive attribute is the binary label in "Gender". In the experiments, we used 8,000 training samples and 4,000 test samples.\\
\item \textbf{Adult} dataset comprises census records with 64 binary features and income labels ($>50K$ or $\leq50K$ annually). Gender serves as the sensitive attribute
\footnote{https://archive.ics.uci.edu/dataset/2/adult}. \\
We utilize 12,000 samples for training and 3,000 for testing.
\item \textbf{COMPAS} dataset contains criminal recidivism records with 12 features and binary labels indicating two-year recidivism outcomes. We use race (Caucasian/Non-caucasian) as the sensitive attribute
\footnote{https://github.com/propublica/compas-analysis}
. The dataset is split into 3,000 training and 600 test samples.\\

\end{enumerate}

\item \textbf{Distributed Settings.} For the Adult dataset, we create a five-client federation where Client 1 ("Underrepresented") receives 2,000 male and 400 female samples, while Clients 2-5 ("Overrepresented") each obtain 400 male and 2,000 female samples. For COMPAS, following the same pattern, Client 1 is allocated 600 Caucasian and 100 Non-caucasian samples, with Clients 2-5 each receiving 100 Caucasian and 600 Non-caucasian samples. \\

We distribute CelebA and UTKFace datasets across ten clients with controlled demographic imbalances. For CelebA, Clients 1-2 ("Underrepresented") receive 200 male and 1,000 female samples each, while Clients 3-10 ("Overrepresented") receive 1,000 male and 200 female samples each. Similarly for UTKFace, Minority clients are allocated 100 male and 700 female samples each, with Majority clients receiving the inverse distribution.\\

This partitioning results in global training sets with the following distributions: Adult (8,400 Female, 3,600 Male), COMPAS (2,100 Non-caucasian, 900 Caucasian), CelebA (8,400 Male, 3,600 Female), and UTKFace (5,800 Male, 2,200 Female). All test sets preserve identical sensitive attribute proportions to their respective training sets.\\

To simulate realistic heterogeneity while preserving the overall average proportions, we introduce slight variations using Dirichlet distribution sampling. Specifically, we employ different $\alpha$ values for minority and majority clients, modeling the data as a mixture of two distinct Dirichlet distributions. The $\alpha$ values are set to 0.5 and 1 for the two groups, ensuring the sensitive attribute proportions are approximately 2:8 and 8:2, respectively.\\

\item \textbf{Baselines.} KDE \cite{cho2020bfair} is an in-processing fairness algorithm leveraging kernel density estimation, where KDE(local) trains on local datasets independently. Meanwhile, FedAvg+KDE combines model aggregation with local KDE training and pFedMe+KDE combines personalized FL with local KDE training. \\

For tabular datasets, we have: (1) MMPF \cite{martinez2020minimax} employs minimax optimization to enhance worst-case fairness performance across groups. (2) FA \cite{du2021fairness} introduces a kernel-based agnostic federated learning framework addressing fairness constraints. (3) FCFL \cite{cui2021addressing} implements constrained multi-objective optimization at client level.\\

For image datasets, we compare our approach with several state-of-the-art methods: (1) IRM \cite{rosenfeld2020risks}, which establishes a learning framework to identify and leverage invariant correlations across diverse training distributions; (2) RES \cite{romano2020achieving}, which addresses data heterogeneity through strategic resampling techniques; (3) DRO \cite{levy2020large}, which optimizes model performance for worst-case scenarios across different subgroups to enhance fairness; and (4) DiGA \cite{zhang2024distributionally}, which leverages generative modeling to construct unbiased training datasets for fair classifier training.\\

\item \textbf{Model Architecture.} For fair comparison, we implemented identical neural network architectures across all methods for tabular datasets: (1) for the Adult dataset, a multi-layer perceptron (MLP) with 4 hidden layers and 256 neurons per layer, (2) for the COMPAS dataset, an MLP with 2 hidden layers and 64 neurons per layer, (3) for the CelebA and UTKFace, our proposed method applies DINOv2 (ViT-B/14) and CLIP (ViT-B/16) for feature extraction, followed by a single linear layer with 768 and 512 neurons, respectively. The self-implemented CNN uses 3 convolutional layers, while MLP also uses a 3-layer structure. All the experiments are done on a single Nvidia GeForce RTX 3090.\\

\begin{table*}[htbp]\Large
    \centering
    \renewcommand{\arraystretch}{1.25}
    \resizebox{0.9\textwidth}{!}{
    \tabcolsep=0.5cm
    \begin{tabular}{lcccccccc}
    \toprule
    & \multicolumn{2}{c}{\textbf{Minority}} & \multicolumn{2}{c}{\textbf{Majority}} & \multicolumn{2}{c}{\textbf{Worst}} & \multicolumn{2}{c}{\textbf{Average}}\\
    \cmidrule(l){2-3} 
    \cmidrule(l){4-5}
    \cmidrule(l){6-7}
    \cmidrule(l){8-9}
    & \textbf{Acc($\uparrow$)} & \textbf{DDP($\downarrow$)} & \textbf{Acc($\uparrow$)} & \textbf{DDP($\downarrow$)} & \textbf{Acc($\uparrow$)} & \textbf{DDP($\downarrow$)}
     & \textbf{Acc($\uparrow$)} & \textbf{DDP($\downarrow$)}\\
    \midrule
    
    KDE(local) (20 clients) & 75.2\% & \textbf{0.035} & 78.5\% & 0.028 & 73.5\% & \textbf{0.045} & 77.0\% & 0.031\\
    pFedMe+KDE (20 clients) & 78.1\% & 0.042 & 83.5\% & 0.032 & 77.3\% & 0.057 & 82.4\% & 0.036\\
    pFedFair (20 clients) & \textbf{82.5\%} & 0.046 & \textbf{86.8\%} & \textbf{0.024} & \textbf{82.0\%} & 0.056 & \textbf{86.0\%} & \textbf{0.030}\\
    \midrule
    KDE(local) (50 clients) & 74.1\% & \textbf{0.043} & 75.5\% & \textbf{0.028} & 73.7\% & \textbf{0.051} & 75.0\% & \textbf{0.036}\\
    pFedMe+KDE (50 clients) & 77.7\% & 0.056 & 82.0\% & 0.037 & 76.0\% & 
    0.062 & 81.2\% & 0.044\\
    pFedFair (50 clients) & \textbf{81.1\%} & 0.047 & \textbf{85.2\%} & 0.039 & \textbf{80.4\%} & 0.061 & \textbf{84.4\%} & 0.043\\
    \bottomrule
    \end{tabular}}
    \caption{Experimental evaluation on 20 and 50 clients for the Adult dataset, against locally-trained KDE and pFedMe+KDE baselines. All clients maintain controlled sensitive attribute distributions with a fixed binary ratio ($20\%$:$80\%$) across individual clients.}
    \label{pfl_fix}

    \vspace{1cm}
    \centering
    \renewcommand{\arraystretch}{1.25}
    \resizebox{0.9\textwidth}{!}{
    \tabcolsep=0.5cm
    \begin{tabular}{lcccccccc}
    \toprule
    & \multicolumn{2}{c}{\textbf{Minority}} & \multicolumn{2}{c}{\textbf{Majority}} & \multicolumn{2}{c}{\textbf{Worst}} & \multicolumn{2}{c}{\textbf{Average}}\\
    \cmidrule(l){2-3} 
    \cmidrule(l){4-5}
    \cmidrule(l){6-7}
    \cmidrule(l){8-9}
    & Acc($\uparrow$) & DDP($\downarrow$) & Acc($\uparrow$) & DDP($\downarrow$) & Acc($\uparrow$) & DDP($\downarrow$)
     & Acc($\uparrow$) & DDP($\downarrow$)\\
    \midrule
    KDE(local) (20 clients) & 75.8\% & \textbf{0.040} & 78.0\% & 0.032 & 72.2\% & \textbf{0.048} & 76.6\% & \textbf{0.036}\\
    pFedMe+KDE (20 clients) & 78.9\% & 0.045 & 83.3\% & 0.039 & 75.3\% & 0.060 & 82.4\% & 0.043\\
    pFedFair (20 clients) & \textbf{83.0\%} & 0.048 & \textbf{86.2\%} & \textbf{0.030} & \textbf{81.1\%} & 0.059 & \textbf{85.5\%} & 0.038\\
    \midrule
    KDE(local) (50 clients) & 73.3\% & \textbf{0.049} & 74.8\% & \textbf{0.036} & 72.1\% & \textbf{0.051} & 74.5\% & \textbf{0.042}\\
    pFedMe+KDE (50 clients) & 77.4\% & 0.062 & 81.5\% & 0.041 & 73.7\% & 
    0.068 & 80.6\% & 0.052\\
    pFedFair (50 clients) & \textbf{81.3\%} & 0.059 & \textbf{84.3\%} & 0.044 & \textbf{79.2\%} & 0.059 & \textbf{83.7\%} & 0.051\\
    \bottomrule
    \end{tabular}}
    \caption{Experimental evaluation on 20 clients and 50 clients for the Adult dataset. Minority and majority clients have different sensitive attribute distributions sampled from two different Direchlet distributions, shown in Figure \ref{fig:direchlet}.}
    \label{pfl_direchlet}
\end{table*}

\item \textbf{Hyper-parameters.} The hyperparameters of pFedFair are carefully tuned based on the characteristics of different datasets. Specifically, the parameter $\lambda$, which controls the balance between global model consistency and local fairness-aware optimization, is set to 0.4 to achieve optimal performance trade-off. To explore various levels of fairness constraints, the fairness regularization parameter $\gamma$ ranges from 0 to 0.9. For the optimization schedule, we set the number of outer loops to 1, while the number of inner loops varies by data type: 10 for tabular datasets and 3 for image datasets. The learning rate for outer loop optimization is configured to be 5-10 times larger than that of the inner loop to ensure stable convergence.
\end{itemize}

\subsection{Results for more clients on Adult and COMPAS}
\label{appendix:more}

We also present results comparing KDE (local), pFedMe+KDE, and pFedFair for 20 and 50 clients on the Adult dataset. In Table~\ref{pfl_fix}, we use fixed sensitive attribute proportions of 2:8 for minority clients and 8:2 for majority clients. In Table~\ref{pfl_direchlet}, we sample varying proportions for majority and minority clients using the Dirichlet distribution, modeled as a mixture of two Dirichlet distributions with different $\alpha$ values. In both scenarios, pFedFair achieves the best fairness-accuracy trade-off compared to other baselines, demonstrating its robustness even with limited data.

\begin{figure*}[htbp]
    \centering
    \subfloat[20 clients sensitive attribute distribution]{%
        \includegraphics[width=0.44\textwidth]{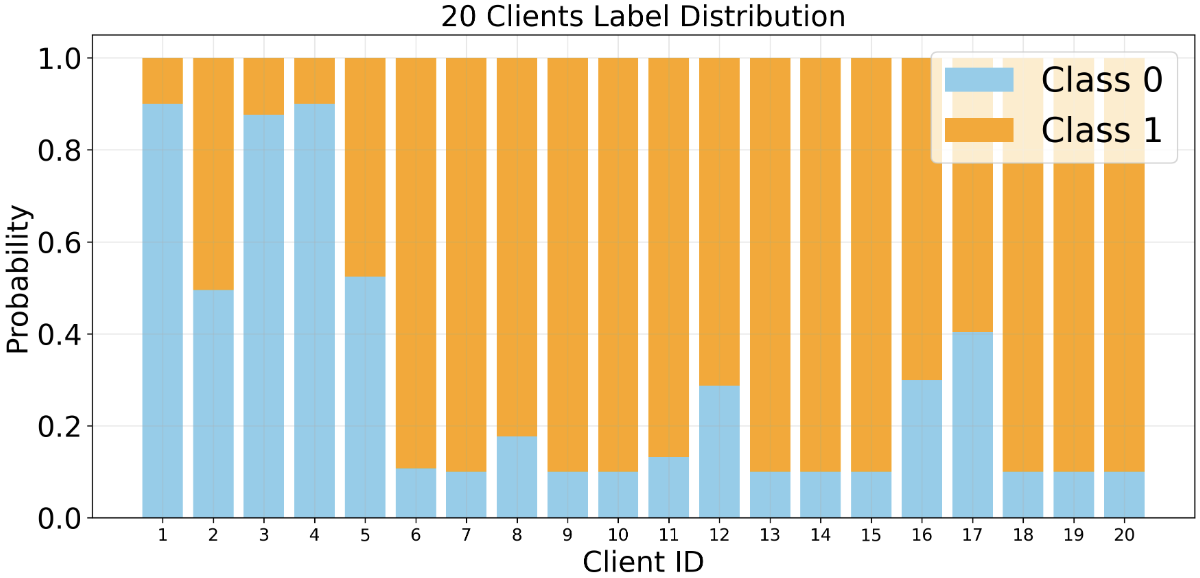}}%
    \subfloat[50 clients sensitive attribute distribution]{%
        \includegraphics[width=0.55\textwidth]{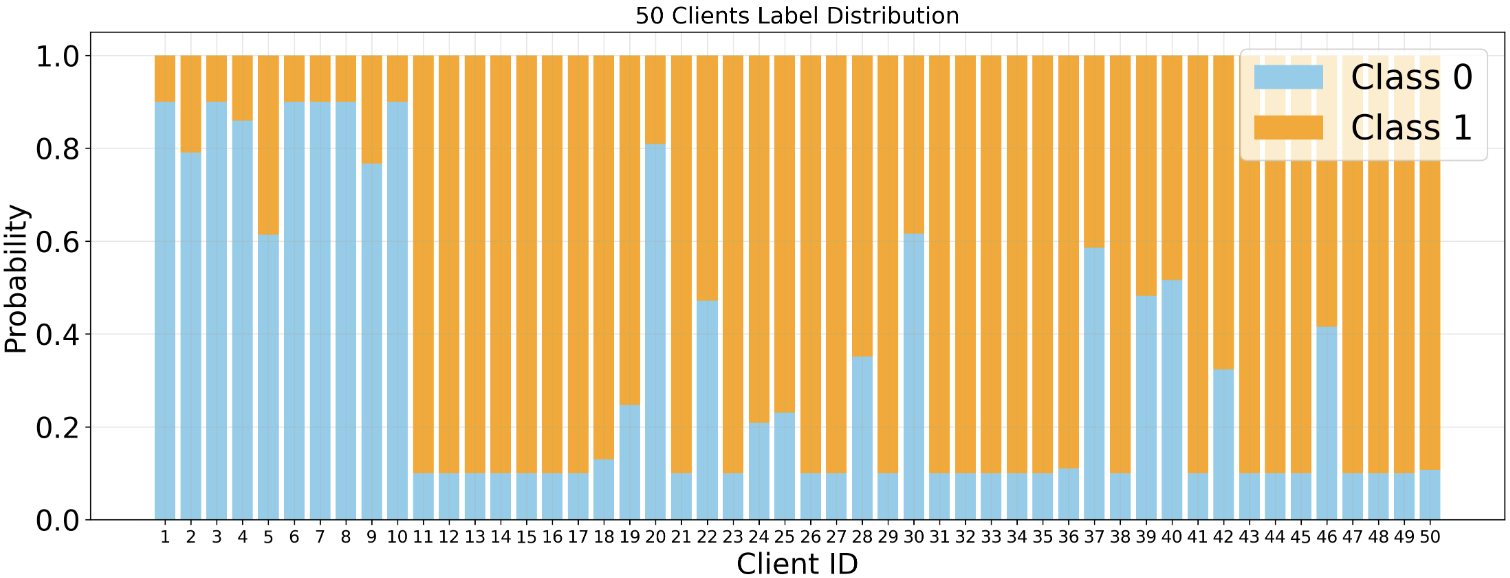}}%
    \caption{Client-level sensitive attribute distributions are sampled from a mixture of two Dirichlet distributions, ensuring heterogeneous data partitioning between minority clients (the initial $20\%$ of the client population) and majority clients (the remaining $80\%$).}
    \label{fig:direchlet}
\end{figure*}

\begin{table}[t]\large
    \centering
    \renewcommand{\arraystretch}{1.25}
    \resizebox{0.66\textwidth}{!}{
    \begin{tabular}{lcc}
    \toprule
    \textbf{Models} & \textbf{Training time($\downarrow$)} & \textbf{Model Parameters($\downarrow$)}\\
    \midrule
    ResNet-18 & 331.2s & 11.71M\\
    ViT-B/16 & 557.7s & 86.45M\\
    CNN  & 278.5s & 3.22M\\
    MLP  & 195.4s & 1.67M\\
    \cmidrule{1-3}
    \textbf{Fair Linear (DINOv2)} & \textbf{15.3s} & \textbf{0.0008M}\\
    \textbf{Fair Linear (CLIP)} & \textbf{13.7s} & \textbf{0.0005M}\\
    \bottomrule
    \end{tabular}}
    \caption{Comparison of training time (for each epoch) and model parameters in centralized CelebA dataset.}
    \label{train_celeba}
\end{table}

\begin{table}[t]\Large
    \centering
    \renewcommand{\arraystretch}{1.25}
    \resizebox{0.66\textwidth}{!}{
    \tabcolsep=0.5cm
    \begin{tabular}{lcccc}
    \toprule
    & \multicolumn{2}{c}{\textbf{CelebA}} & \multicolumn{2}{c}{\textbf{UTKFace}}\\
    \cmidrule(l){2-3} 
    \cmidrule(l){4-5}
    \textbf{Methods} & \textbf{Acc($\uparrow$)} & \textbf{DDP($\downarrow$)} & \textbf{Acc($\uparrow$)} & \textbf{DDP($\downarrow$)}\\
    \midrule
    ERM & 90.2\% & 0.192 & 90.6\% & 0.101\\
    \midrule
    IRM \cite{rosenfeld2020risks} & 87.2\% & 0.063 & 88.1\% & 0.050\\
    RES \cite{romano2020achieving}& 88.0\% & 0.059 & 88.6\% & 0.045 \\
    DRO \cite{levy2020large}& 87.7\% & 0.098 & 88.9\% & 0.067\\
    DiGA \cite{zhang2024distributionally}& 88.1\% & 0.046 & \textbf{89.3\%} & 0.036\\
    \textbf{Fair Linear (DINOv2)} & \textbf{88.4\%} & \textbf{0.035} & 89.0\% & \textbf{0.028}\\
    \bottomrule
    \end{tabular}}
    \caption{Performance comparison of fair facial recognition methods in centralized settings on CelebA and UTKFace datasets. All methods are optimized to minimize DDP while maintaining comparable accuracy levels.}
    \label{central_baseline}
\end{table}

\subsection{Results in centralized fair visual recognition}
\label{appendix:central}

As shown in Table~\ref{train_celeba}, our approach significantly reduces computational overhead in both training time and model parameters. The proposed methodology outperforms baseline architectures and offers flexibility in incorporating various fairness estimators, establishing a more efficient approach for fair visual recognition. Additionally, Table.~\ref{central_baseline} presents a comparative analysis between Fair Linear and other state-of-the-art fair facial recognition methods. Our approach demonstrates competitive performance in centralized settings while offering computational efficiency advantages.

\subsection{Additional Results in Heterogeneous FL Settings}

\begin{figure*}[htbp]
    \centering
    \subfloat[Average Performance on CelebA)]{%
        \includegraphics[width=0.35\textwidth]{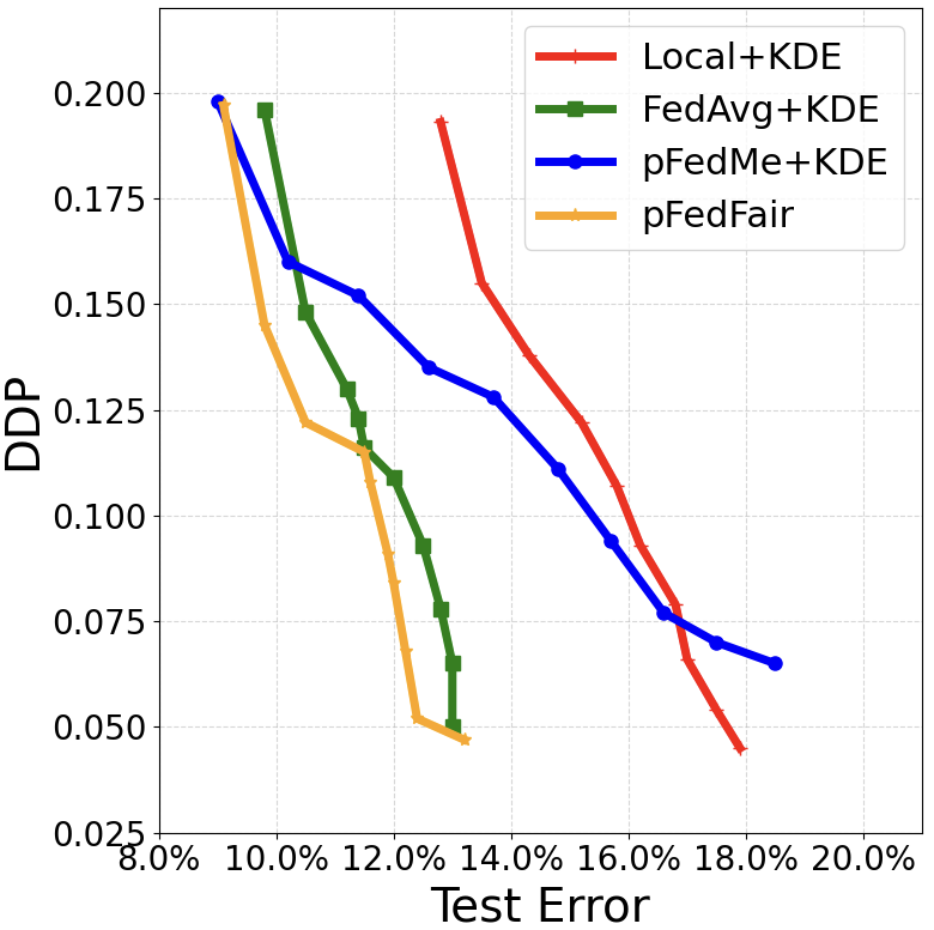}}%
    \subfloat[Average Performance on UTKFace)]{%
        \includegraphics[width=0.35\textwidth]{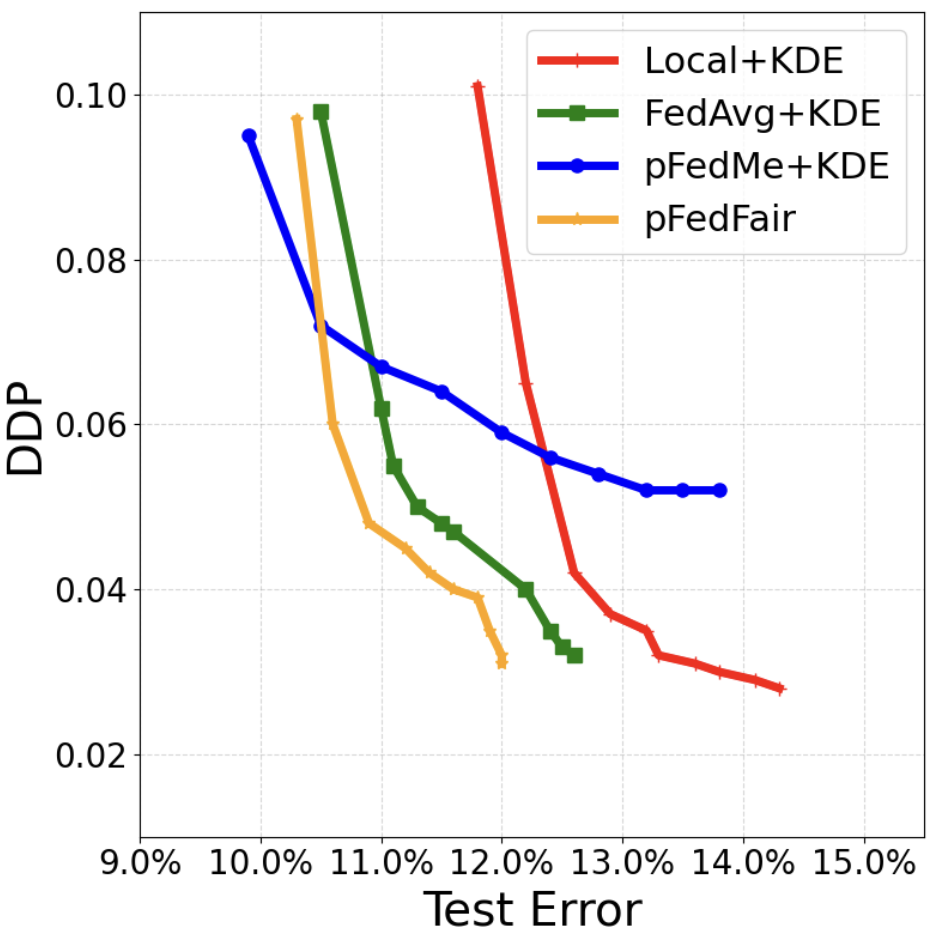}}%
    \caption{Experimental results on the average performance trade-off between Test Error ($\downarrow$) and DDP ($\downarrow$) on the CelebA and UTKFace datasets.}
    \label{fig:average}
\end{figure*}

We compare the performance of pFedFair against three baselines—KDE (local), FedAvg+KDE, and pFedMe+KDE—by evaluating the average performance across all clients on the CelebA and UTKFace datasets (see Fig.~\ref{fig:average}). Among these, FedAvg+KDE aims to optimize toward a global objective, often favoring the majority data distribution. In contrast, pFedFair not only slightly outperforms FedAvg+KDE in terms of overall performance but also significantly exceeds the results of the other two baselines, KDE (local) and pFedMe+KDE. These results highlight the effectiveness of pFedFair in exploring client-level optimal fairness-accuracy trade-offs. Importantly, it achieves this balance without causing substantial harm to the overall performance across the entire dataset.

\end{appendices}
\end{document}